\pdfoutput=1

\PassOptionsToPackage{table,xcdraw}{xcolor}

\documentclass[11pt]{article}

\usepackage[final]{acl}   
\usepackage{fdsymbol}

\usepackage{times}
\usepackage{latexsym}
\usepackage{makecell}
\usepackage{tablefootnote}
\usepackage{longtable}
\usepackage{adjustbox}
\usepackage{enumitem}
\usepackage[T1]{fontenc}

\usepackage[utf8]{inputenc}

\usepackage{microtype}

\usepackage{hyperref}
\usepackage{booktabs}
\usepackage{graphicx}
\usepackage[table,xcdraw]{xcolor}
\definecolor{orange}{RGB}{237,125,49}
\definecolor{green}{RGB}{112,173,71}
\definecolor{blue}{RGB}{68,114,196}
\definecolor{red}{RGB}{255,0,0}
\definecolor{purple}{RGB}{112,48,160}
\definecolor{brown}{RGB}{165,42,42}
\definecolor{gold}{rgb}{0.83, 0.69, 0.22}
\definecolor{fluorescentpink}{rgb}{1.0, 0.08, 0.58}
\definecolor{lightseagreen}{rgb}{0.13, 0.7, 0.67}
\definecolor{darkpastelgreen}{rgb}{0.01, 0.75, 0.24}
\graphicspath{{./imgs/}}
\usepackage{footmisc}

\usepackage{floatrow}
\usepackage{microtype}

\usepackage{caption}
\usepackage{subcaption}
\usepackage{array, makecell} %

\usepackage{pifont}

\usepackage{tabularx,colortbl}

\usepackage{tikz}
\usepackage{collcell}

\usepackage{etoolbox}

\newcolumntype{?}{!{\vrule width 1.5pt}}

\newtoggle{inTableHeader}
\toggletrue{inTableHeader}
\newcommand*{\StartTableHeader}{\global\toggletrue{inTableHeader}}%
\let\OldTabular\tabular%
\let\OldEndTabular\endtabular%
\renewenvironment{tabular}{\StartTableHeader\OldTabular}{\OldEndTabular\StartTableHeader}%

\newcommand*{\MinNumber}{-1.0}%
\newcommand*{\MidNumber}{0.0} %
\newcommand*{\MaxNumber}{1.0}%

\newcommand{\ApplyGradient}[1]{%
  \iftoggle{inTableHeader}{#1}{
    \ifdim #1 pt > \MidNumber pt
        \pgfmathsetmacro{\PercentColor}{max(min(100.0*(#1 - \MidNumber)/(\MaxNumber-\MidNumber),100.0),0.00)} %
        \hspace{-0.33em}\colorbox{yellow!\PercentColor!blue}{#1}
    \else
        \pgfmathsetmacro{\PercentColor}{max(min(100.0*(\MidNumber - #1)/(\MidNumber-\MinNumber),100.0),0.00)} %
        \hspace{-0.33em}\colorbox{blue!\PercentColor!blue}{#1}
    \fi
  }}
\newcolumntype{R}{>{\collectcell\ApplyGradient}c<{\endcollectcell}}

\usepackage{amsmath}
\usepackage{amsfonts,bm}
\usepackage{xspace}

\definecolor{mypink3}{cmyk}{0, 0.7808, 0.4429, 0.1412}

\makeatletter   
\newcommand{\sveryshortarrow}[1][3pt]{\mathrel{%
    \vcenter{\hbox{\rule[-.5\fontdimen8\scriptfont3]
               {\scriptratio\dimexpr#1\relax}{\fontdimen8\scriptfont3}}}%
   \mkern-4mu\hbox{\let\f@size\sf@size\usefont{U}{lasy}{m}{n}\symbol{41}}}}
\makeatother

\def\eqref#1{equation~\ref{#1}}

\def\1{\bm{1}}

\def\m1{{\bm{1}}}

\DeclareMathAlphabet{\mathsfit}{\encodingdefault}{\sfdefault}{m}{sl}
\SetMathAlphabet{\mathsfit}{bold}{\encodingdefault}{\sfdefault}{bx}{n}

\usepackage[nameinlink]{cleveref}
\crefformat{section}{\S#2#1#3} 
\crefname{algorithm}{Alg.}{Algs.}
\crefformat{subsection}{\S#2#1#3}
\Crefname{equation}{Eq.}{Eqs.}
\Crefname{figure}{Fig.}{Figs.}

\usepackage[colorinlistoftodos,prependcaption,textsize=tiny]{todonotes}

\newcommand{\change}[1]{{\leavevmode\color{black}#1}}

\usepackage{soul}

\usepackage{float}
\definecolor{azure}{rgb}{0.0, 0.5, 1.0}

\usepackage{multirow}
\usepackage{hhline}

\usepackage{arydshln} 
\usepackage[group-separator={,}]{siunitx} 

\title{Are Large Vision Language Models up to the Challenge of \\ Chart Comprehension and Reasoning?}

\author{
Mohammed Saidul Islam$^{\clubsuit}$
, \ Raian Rahman$^{\spadesuit}$
, \ Ahmed Masry$^{\clubsuit}$\thanks{\ \ Equal contribution.}, \\
{\bf \ Md Tahmid Rahman Laskar$^{\clubsuit\heartsuit}$}\footnotemark[1]
{\bf, \ Mir Tafseer Nayeem$^{\vardiamondsuit}$}\footnotemark[1] 
{\bf, \ Enamul Hoque$^{\clubsuit}$} \\
$^\clubsuit$York University, Canada,
$^\spadesuit$Islamic University of Technology, Bangladesh, \\ $^\vardiamondsuit$University of Alberta, Canada, $^\heartsuit$Dialpad Inc., Canada \\
\{saidulis, masry20, tahmid20, enamulh\}@yorku.ca, raianrahman@iut-dhaka.edu, \\
mnayeem@ualberta.ca
}
\vspace{1cm}

\begin{document}
\maketitle

\begin{abstract} 
Natural language is a powerful complementary modality of communication for data visualizations, such as bar and line charts.  To facilitate chart-based reasoning using natural language, various downstream tasks have been introduced recently such as chart question answering, chart summarization, and fact-checking with charts. These tasks pose a unique challenge, demanding both vision-language reasoning and a nuanced understanding of chart data tables, visual encodings, and natural language instructions. Despite the recent success of Large Language Models (LLMs) across diverse NLP tasks, their abilities and limitations in the realm of data visualization remain under-explored, possibly due to their lack of multi-modal capabilities. To bridge the gap, this paper presents one of the first comprehensive evaluations of the recently developed large vision language models (LVLMs) 
for chart understanding and reasoning tasks. Our evaluation includes a comprehensive assessment of both closed and open-sourced LVLMs
across five major chart reasoning tasks. Furthermore, we perform a qualitative evaluation of LVLMs' performance on a diverse range of charts, aiming to provide a thorough analysis. 
Our findings reveal that while LVLMs demonstrate impressive abilities in generating fluent texts covering high-level data insights, they also encounter common problems like hallucinations, factual errors, and data bias. We highlight the key strengths and limitations of LVLMs in chart comprehension tasks, offering insights for future research 
\footnote{We make all our prompts as well as LVLMs' responses open source \href{https://github.com/saidul-islam98/LVLM-ChartEval}{here}.}.
\end{abstract}

\begin{figure}[t]
    \centering
    \includegraphics[width=0.95\columnwidth]{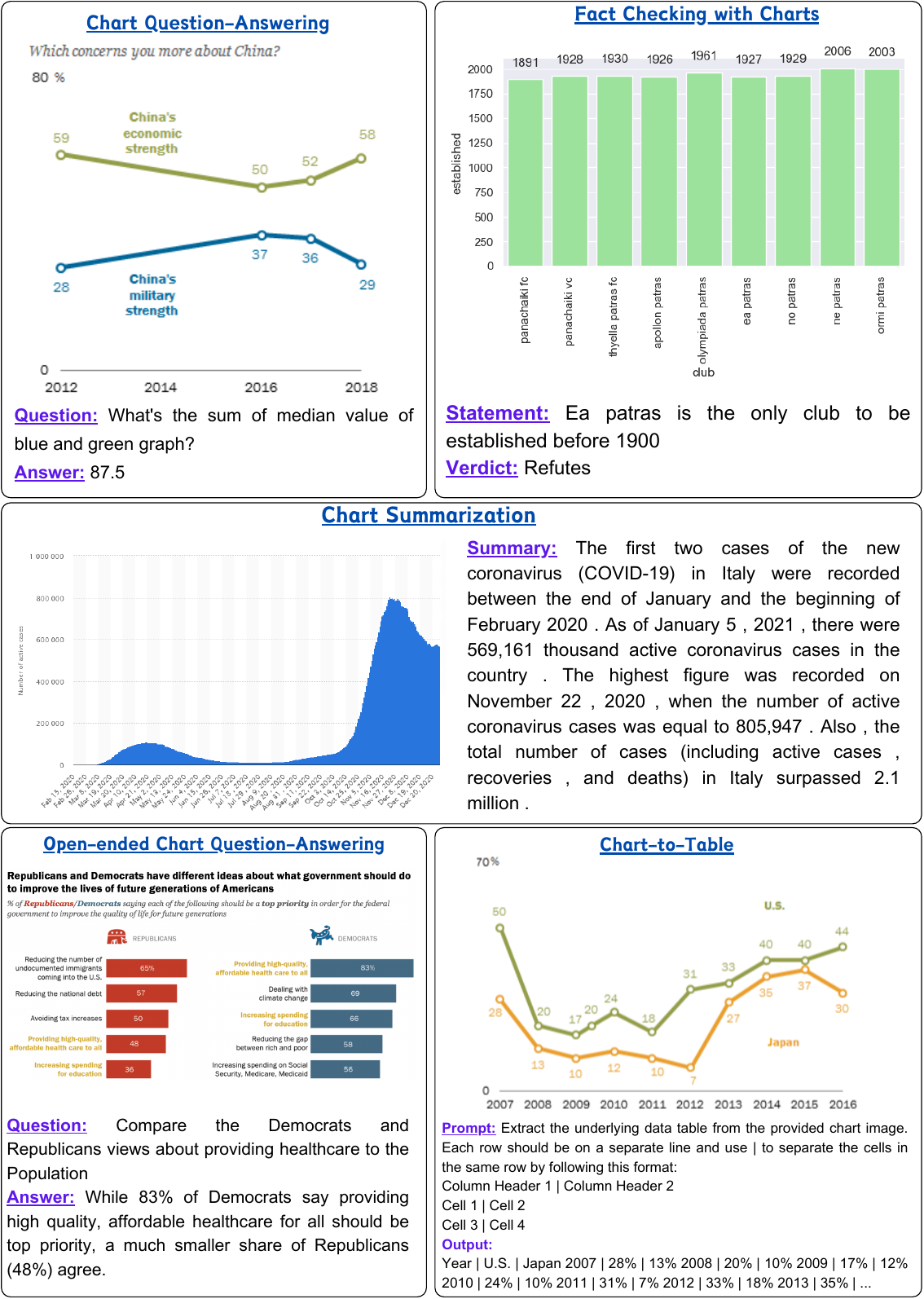}
    \caption{Chart comprehension and reasoning tasks.}
    \label{fig:tasks}
\end{figure}

\section{Introduction}
\label{sec-intro}
Natural language and visualizations are two powerful complementary modalities to communicate data insights. While visualizations can be very effective in finding patterns, trends, and outliers in data, natural language can help explain the key insights in visualizations and answer questions about data~\cite{hoque2022chartSurvey}. People commonly integrate text with graphical charts as this combination helps direct attention to specific elements of the chart and offers necessary explanations that might otherwise go unnoticed \cite{stokes2022striking}. Moreover, interfaces that use natural language to interact with charts have other 
benefits, such as, enhancing chart accessibility~\cite{seechart-2023} and supporting visualization education~\cite{bach2023challenges}. 

Given the importance of chart comprehension and reasoning, researchers have introduced various tasks for the development of automated methods to aid users in chart analysis (see \Cref{fig:tasks}). These include chart question answering~\cite{masry-etal-2022-chartqa, open-CQA, lee2022pix2struct}, natural language generation for charts \cite{obeid-hoque-2020-chart, chart-to-text-acl}, and fact-checking with charts \cite{akhtar-etal-2023-reading, akhtar2023chartcheck}. To build automated systems for these tasks, a prevalent approach involves pre-training of models 
\cite{liu2022matcha, masry-etal-2023-unichart} and development of frameworks without heuristic rules \cite{cheng2023chartreader} on language and vision tasks~\cite{ijcai2022p762}. However, in recent years, there has been dramatic progress in the development and widespread adoption of LLMs \cite{anil2023palm, chowdhery2023palm,openai2023gpt4, touvron2023llama, touvron2023llama2}. 
While in the beginning, the LLMs were only capable of processing textual data, the rapid progress in this field has paved the way for the development of multimodal LLMs (in other words, LVLMs),  such as GPT-4V \cite{openai2023gpt4}, Gemini \cite{geminiteam2023gemini}, Claude-3 \cite{Claude},
Phi-3 \cite{abdin2024phi3},  LLaVA \cite{liu2023visual}, and MiniGPT-4 \cite{zhu2023minigpt4}. Given the rapid rise of these LVLMs, there is a pressing question: \emph{Are LVLMs up to the challenge of chart
comprehension and reasoning?}

In this paper, we aim to answer this question by investigating the capabilities and limitations of LVLMs in the chart reasoning and comprehension domain. Specifically, we examine whether the latest state-of-the-art (SoTA) LVLMs can effectively interpret charts as well as identify key insights 
solely based on the chart images. This setup is crucial in real-world scenarios where the underlying data tables of charts are often unavailable. To this end, we performed extensive qualitative and quantitative analyses of the performance of LVLMs on five downstream tasks across \emph{seven} benchmark datasets: \textbf{ChartQA} \cite{masry-etal-2022-chartqa}, \textbf{OpenCQA}~\cite{kantharaj2022opencqa}, \textbf{Chart Summarization} (Chart-to-Text \cite{kantharaj-etal-2022-chart}, and \textbf{Vistext} \cite{2023-vistext}), \textbf{Fact-checking} (ChartFC \cite{akhtar-etal-2023-reading}, \textbf{ChartCheck }\cite{akhtar2023chartcheck}), and \textbf{Chart-to-Table} \cite{Choi2019VisualizingFT}. 

Specifically, this work presents the first detailed analyses of LVLMs on seven chart domain benchmarks, making the following main contributions:

\vspace{7pt}

\textbf{(1)} Existing SoTA models typically report quantitative performance on ChartQA without a detailed analysis of their capabilities and limitations. We examine LVLMs' performance using advanced techniques like Chain-of-Thought \cite{wei2023chainofthought} and Program-aided Language Models \cite{gao2023pal} (\Cref{subsec-reschartqa}).

\vspace{7pt}

\textbf{(2)} Unlike most closed-source models that focus only on factoid question answering (ChartQA), we evaluate LVLMs on other important tasks such as OpenCQA and Chart Summarization, presenting the first analysis of LVLMs' capability in generating open-ended responses (\Cref{subsec-rescharttotext}, \Cref{subsec-resopencqa}).

\vspace{7pt}

\textbf{(3)} Hallucinations, factual errors, and bias are common issues for many LVLMs. We investigate these problems through various analyses (\Cref{subsec-resfactcheck}, \Cref{subsec-c2t-hallucination-analysis} and \Cref{sec:bias}), including the adoption of an error taxonomy \cite{mishra2024finegrained} for hallucinations. 

\vspace{7pt}

\textbf{(4)} We address the fundamental question of how effectively LVLMs can interpret charts by measuring their ability to extract data from chart images, being the first to thoroughly examine this (\Cref{subsec-charttotable}).

\vspace{7pt}

\textbf{(5)} Text generation tasks require models to describe high-level trends and outliers, as well as low-level chart details like colors. We analyze how often and how accurately models cover different types of semantic content using the 4-level framework \cite{lundgard2021accessible} (\Cref{subsec-semantics}).

\section{Related Work}
\label{sec-relwork}

\paragraph{Chart-related Downstream Tasks:}
\label{subsec-charttasks} 
Several downstream tasks associated with charts have been proposed recently. Chart Question Answering refers to answering factoid questions 
regarding charts \cite{dvqa, figureqa, plotqa, masry-etal-2022-chartqa, xu2024chartbench}. 
In contrast, Open-ended Question Answering (OpenCQA) requires explanatory responses by interpreting chart data \cite{open-CQA}. The Chart Summarization task \cite{chart-to-text-acl, obeid-hoque-2020-chart, 2023-vistext, Rahman_2023} involves creating natural language descriptions from charts,   
Chart-to-Table focuses on converting charts into data tables \cite{Choi2019VisualizingFT}, and Fact-checking with charts \cite{akhtar-etal-2023-reading, akhtar2023chartcheck} focuses on verifying factual statements related to charts. 
While there are other areas like infographic comprehension \cite{mathew2022infographicvqa} and science diagram QA
\cite{kembhavi2016diagram}, this study is 
devoted to chart-related tasks. 

\paragraph{Evaluation of LVLMs:}
\label{subsec-llmeval}
OpenAI's introduction of GPT-4V marked a significant advancement, outperforming other LVLMs proposed in \cite{liu2023visual, dai2023instructblip, zhu2023minigpt4}, particularly in scenarios data-scarce scenarios. Google's Gemini \cite{geminiteam2023gemini} and Anthropic's Claude-3 \cite{Claude} have recently emerged as strong competitors, and Microsoft's open-source Phi-3 model achieved performance comparable to closed-source LVLMs in multimodal tasks \cite{abdin2024phi3}.
While some studies compared Gemini and GPT-4V models on image recognition and understanding tasks \cite{qi2023gemini, fu2023challenger}, other works have focused on solving chart-related tasks only using data tables \cite{do2023llms,huang2023lvlms}.  
Additionally, some recent studies have proposed a benchmark dataset \cite{guan2024hallusionbench} to study image-context reasoning, introduced a new LLM for improved open-ended visual question-answering \cite{hu2023bliva}, and assessed GPT-4V-type models on tasks requiring structured reasoning \cite{singh2023assessing}.
However, these studies address only one chart-related task (i.e., Chart Question Answering) with quantitative analysis, whereas our work presents the most comprehensive evaluation of LVLMs in the chart reasoning and comprehension domain with five downstream tasks with a wider range of qualitative and quantitative analyses. Therefore, our work clearly distinguishes itself from the abovementioned works.

\begin{table*}[ht]
\scriptsize
 \setlength\extrarowheight{2pt}
 \centering
    \scalebox{0.83}{\begin{tabular}{cccccc|cc|cccccc|cc|cc|cc|cc|cc}  
  \toprule
  
  \multicolumn{8}{c|}{\textbf{ChartQA}}  & \multicolumn{6}{c|}{\textbf{Chart-to-Table}} & \multicolumn{2}{c|}{\textbf{OpenCQA}}  & \multicolumn{2}{c|}{\textbf{Chart-to-Text}} & \multicolumn{2}{c|}{\textbf{Vistext}} & \multicolumn{2}{c|}{\textbf{ChartFC}} & \multicolumn{2}{c}{\textbf{ChartCheck}}   \\  \hline \midrule
  \multicolumn{2}{c}{Human} & \multicolumn{2}{c}{Augmented} & \multicolumn{2}{c|}{Total} & \multicolumn{2}{c|}{ChartQA$^*$} & \multicolumn{2}{c}{Human} & \multicolumn{2}{c}{Augmented} & \multicolumn{2}{c|}{Total} & \multicolumn {1}{c}{} & \multicolumn {1}{c}{} & \multicolumn {1}{c}{} & \multicolumn {1}{c}{} & \multicolumn {1}{c}{} & \multicolumn {1}{c}{} & \multicolumn {1}{c}{} & \\  
  \cmidrule(lr){1-2} \cmidrule(lr){3-4} \cmidrule(lr){5-6} \cmidrule(lr){7-8} \cmidrule(lr){9-10} \cmidrule(lr){11-12} \cmidrule(lr){13-14} \cmidrule(lr){15-16} \cmidrule(lr){17-18}\cmidrule(lr){19-20}\cmidrule(lr){21-22}\cmidrule(lr){23-24}
  
  Charts & Qs. & Charts & Qs. & Charts & Qs. & Charts & Qs. & Charts & Tables. & Charts & Tables. & Charts & Tables. & Charts & Qs. & Pew & Stat. & Chart & Summ. & Supp. & Ref. & Test1 & Test2  \\
                                             
  \midrule
  625 & 1250 & 987 & 1250 & 1612 & 2500 & 1340 & 2192 & 625 & 625 & 987 & 987 & 1612 & 1612 & 1159 & 1159 & 1393 & 5222 & 882 & 1270 & 885 & 706 & 937 & 981 \\ \hline
  
 \end{tabular}}
\caption{
Test set of seven benchmarks:
Here, ``Qs.'', ``Pew'', and ``Stat.'' refer to Questions, Pew charts, and Statista charts, respectively. ``Supp.'' and ``Ref.'' denote the Support and Refute classes in ChartFC. 
ChartQA$^*$ denotes charts from the ChartQA test set without data labels.
}
 \label{tab:overview-datasets}
\end{table*}

\section{Methodology}
\label{sec-method}
\subsection{Tasks and Datasets}
\label{subsec-tasks}
Since chart comprehension and reasoning is a relatively new topic of research, very few tasks are proposed so far and there is a scarcity of benchmark resources, i.e., very few datasets, and models. Nevertheless, we have included all existing major chart-related downstream tasks for experiments.

\paragraph{(1) Factoid Chart Question Answering:} For this task, we use ChartQA \cite{masry-etal-2022-chartqa}, a popular benchmark 
with a focus on visual and logical reasoning questions and was used as the only dataset for this task by GPT-4V \cite{gpt4-report} and Gemini \cite{gemini-report} in their released reports. It features human-written questions from four real-world sources covering a wide range of topics. 

\paragraph{(2) Chart Summarization:} We choose Chart-to-Text \cite{chart-to-text-acl}, a large-scale benchmark for chart summarization as well as  Vistext \cite{2023-vistext}, another recent chart captioning dataset. 

\paragraph{(3) Open-ended Chart Question Answering:} We consider OpenCQA \cite{open-CQA}, the only  QA benchmark available for this task in which answers are provided as explanatory texts.  

\paragraph{(4) Fact-checking with Charts:} We utilize two currently available datasets: ChartFC dataset which contains (\emph{statement}, \emph{verdict}) as pairs \cite{akhtar-etal-2023-reading} and ChartCheck which has more diverse charts and contain explanations for verdicts. 

\paragraph{(5) Chart-to-Table:} We use the chart-table pairs from the ChartQA test set for the evaluation in this task. We created a new version of ChartQA, named (ChartQA$^*$), by modifying the original charts to exclude explicit data value labels. This setup was introduced to evaluate whether the performance of LVLMs depends on explicit data labels or their ability to interpret data from the visual elements in the charts (more details in \S\ref{app:chartqa-no-label}).
In addition to the above tasks, we evaluate the semantic richness of the model's response by crafting a small dataset of \num{200} question-answer pairs based on four-level semantic frameworks \cite{lundgard2021accessible}. An overview of the test sets of these benchmarks is presented in \Cref{tab:overview-datasets}.

\subsection{Models}
\change{Since closed-source LVLMs currently achieve the best results in zero-shot scenarios in most vision-language benchmarks \cite{geminiteam2023gemini}, we select the following three: GPT-4V (\textit{gpt-4-1106-preview}), Gemini (\textit{gemini-1.0-pro-vision}), and Claude-3 (\textit{claude-3-haiku@20240307}). While most open-source LVLMs underperform compared to closed-source ones, we include the Phi-3 (\textit{phi-3-vision-128k-instruct}) model due to its impressive benchmark results. We compare these models with current SoTA chart-specific models, MathCha \cite{liu2022matcha} and UniChart \cite{masry-etal-2023-unichart}.} Additionally, we assessed other open-source models like mPlug-DOC-owl-1.5 \cite{hu2024mplugdocowl} and LLaVA-1.5 \cite{liu2024improved}, but due to their subpar performance on chart-related tasks, we excluded them from our discussion.

\subsection{Prompt Construction}
In both qualitative and quantitative evaluation, we first create a task instruction \texttt{T} tailored to a specific test sample \texttt{X}. This instruction is then combined with the existing text of the test sample to form a unified prompt \texttt{P}. This prompt \texttt{P} and the Chart image \texttt{C} are provided as input to the respective LVLMs to generate the corresponding response \texttt{R} (see \S\ref{app:prompt-const} for details and \Cref{tab:prompt-table} for example prompts).

\subsection{Evaluation}
\label{subsec-eval}
\change{In addition to evaluating five benchmark chart-related tasks using existing metrics, we conduct specific evaluations on LVLM-generated responses, focusing on hallucination analysis and semantic coverage. Below, we explain our methodology.}
\change{\subsubsection{Task-specific General Evaluation}}
\paragraph{ChartQA:} We perform a comprehensive quantitative evaluation of the LVLMs on ChartQA in two different experimental setups, i.e., zero-shot Chain-of-Thought (CoT) \cite{wei2023chainofthought}, and Program-aided Language Models (PAL) \cite{gao2023pal} inspired by their recent success in various domains.

\paragraph{Chart Summarization \& OpenCQA:} To evaluate the performance of LVLMs in chart summarization and Open-ended Chart Question-Answering tasks, we follow prior work~\cite{kantharaj-etal-2022-chart}, and leverage a suite of automatic evaluation metrics, including BLEU \cite{papineni2002bleu}, CIDEr \cite{vedantam2015cider}, BLEURT \cite{sellam2020bleurt}, BERTScore \cite{Zhang2017}, and Perplexity \cite{pplx}.

\paragraph{Fact Checking with Charts:} For fact-checking, similar to prior work, we conduct a quantitative evaluation in terms of the F1 metric.

\paragraph{Chart-to-Table:} For this task, we conduct a quantitative evaluation by reporting two metrics: the Relative Number Set Similarity (RNSS) \cite{masry-etal-2022-chartqa} and the Relative Mapping Similarity (RMS) \cite{liu-etal-2023-deplot}. 

\change{\subsubsection{Criteria-based Focused Evaluation}}
\paragraph{Hallucination Analysis:}
Hallucinations and factual errors are common in chart-related tasks \cite{kantharaj-etal-2022-chart, kantharaj2022opencqa, 2023-vistext}. Therefore, we examine this issue using the \textbf{FAVA} \cite{mishra2024finegrained}, which automatically detects and categorizes hallucinations in LLM outputs into different types. Although other works have proposed using specialized metrics for specific evaluation criteria, such as QAFactEval \cite{fabbri-etal-2022-qafacteval} and ChartVE \cite{huang2023lvlms}, which are designed to assess factual consistency between text pairs or text-chart pairs, we plan to incorporate these metrics into our analysis as part of future work.

\paragraph{Generating different semantic levels:} To assess the capability of LVLMs in generating texts about charts with rich semantics, we follow the four-level framework from \citet{lundgard2021accessible}:
\textit{Level \textbf{1}} covers low-level information about the chart, i.e., chart type, axes, etc; \textit{Level \textbf{2}} presents statistical and relational aspects such as descriptive statistics and correlations; \textit{Level \textbf{3}} is about \textit{perceptual and cognitive} phenomena describing complex trends, and patterns, and \textit{Level \textbf{4}} provides domain-specific insights such as social and political contexts. In our study, we evaluate the capabilities of LVLMs in their proficiency at
covering these different types of semantic information. We also analyze their accuracy in interpreting questions and explaining answers across these four levels. Our \textit{Level \textbf{1}} semantic evaluation leveraged a collection of \num{40} charts encompassing a variety of types. We design five \textit{Level \textbf{1}} questions to assess core aspects of chart construction. These questions targeted attributes such as channel encoding (how data is represented visually), chart type (bar, line, pie, etc.), and axis labeling ($x$ and $y$). In the case of \textit{Level \textbf{2}}, we design four questions to assess the ability of the models to identify extrema (maxima, minima) and outliers within charts.  
For \textit{Level \textbf{3}}, we include a wider range of \num{100} chart samples, with \num{28} being line charts. Finally, for \textit{Level \textbf{4}}, to evaluate the domain-specific text generation capability of LVLMs, we employ a test set of \num{200} charts.

\section{Results and Discussion}
\label{sec-results}
\subsection{General Observations}
We  present some key  general observations based on our comprehensive evaluation of the LVLMs:

\vspace{6pt}

\noindent{$\bullet$}\,\, Overall, among closed-source models, GPT-4V is the best performer in discriminative Chart reasoning and comprehension tasks, such as factoid chart question-answering and 
chart fact-checking 
while Gemini is better in open-ended generation tasks such as OpenCQA and Chart-to-Text. However, the open-source model Phi-3 achieves the best results on the ChartQA dataset (\Cref{tab:overview-results}).

\vspace{6pt}

\noindent{$\bullet$}\,\, Gemini is a better Chain-of-Thought reasoner, 
while GPT-4V and Claude-3 is 
better at generating code 
to answer questions about charts (Table \ref{tab:overview-results}).

\vspace{6pt}

\noindent{$\bullet$}\,\, When the data values are not annotated in the charts, the performance of different models 
on ChartQA drops drastically (\Cref{chartqavis-automatic}).

\vspace{6pt}

\noindent{$\bullet$}\,\, \texttt{Entity} and \texttt{Relations} are the most frequent types of hallucinations encountered in all closed-sourced
model-generated text (\Cref{tab:fava-c2t-example}).

\vspace{6pt}

\noindent{$\bullet$}\,\, In general, GPT-4V generates longer summaries with chart-specific (Level 1 \& 3) semantic content, while Gemini generates more succinct summaries with statistical and domain-specific information (Level 2 \& 4), and Claude-3 responses fall in between these two models.

\begin{table*}[ht]
\scriptsize
 \setlength\extrarowheight{1pt}
 \centering
 
 \renewcommand{\arraystretch}{1.14} 
    \scalebox{0.75}{\begin{tabular}{lcccccccccccccccc}  
  \toprule
  
  \multirow{3}{*}{} & \multicolumn{3}{c}{\textbf{ChartQA (zero-shot CoT)}} & \multicolumn{3}{c}{\textbf{ChartQA (zero-shot PAL)}} & 
  \textbf{OpenCQA} & \multicolumn{4}{c}{\textbf{Chart Summarization}} & \multicolumn{3}{c}{\textbf{Chart-Fact-checking}} & \multicolumn{2}{c}{\textbf{Chart-to-Table}}   \\ 
  \cmidrule{2-17}
  \multirow{3}{*}{} & \multicolumn{3}{c}{($Accuracy$)} & \multicolumn{3}{c}{ ($Accuracy$)} & \small ($BLEU$) & \multicolumn{4}{c}{\small ($BLEU$)} & \multicolumn{3}{c}{ ($F1-score$)} & ($RNSS$) &  ($RMS$) \\
  \cmidrule(lr){2-4} \cmidrule(lr){5-7} \cmidrule(lr){8-8} \cmidrule(lr){9-12} \cmidrule(lr){13-15} \cmidrule(lr){16-17}
  
   \textbf{Models} & aug. & human & avg. & aug. & human & avg. &  & Pew & Statista & Vistext(L1) & Vistext(L2/L3) & ChartFC & ChartC(T1) & ChartC(T2) & ChartQA & ChartQA \\ \hline

  \midrule

  Human baseline & - & - & - & - & - & - & - & - & - & - & - & - & - & - & - & 95.7\\
  \hdashline
  Gemini \shortcite{geminiteam2023gemini} & \textbf{74.96} & \textbf{70.72} & \textbf{72.84} & 46.08 & 46.08 & 46.08 & \textbf{6.84} & 35.9 & \textbf{25.8} & \textbf{27.4} & \textbf{15.7} & 65.8 & 71.42 & 68.05 & 85.86 & 54.84\\ 

  GPT-4V \shortcite{openai2023gpt4} & 72.64 & 66.32 & 69.48 & 75.44 & \textbf{65.68} & \textbf{70.56} & 3.31 & 28.5 & 18.2 & 18.2 & 11.3 & \textbf{69.6} & \textbf{73.50} & 71.30  & 81.51 & \textbf{61.97}\\

  Claude-3-haiku \shortcite{Claude} & 47.12 & 42.00 & 44.56 & 76.88 & 63.44 & 70.16 & 4.58 & \textbf{36.9} & \textbf{25.8} & 25.2 & 14.2 & 61.4 & 71.70 & \textbf{73.14} &  \textbf{95.83} & 50.65 \\ 
  
  Phi-3-vision-128k-inst \shortcite{abdin2024phi3} & - & - & \textbf{81.40} & - & - & - & 3.95 & 28.6 & 19.9 & 20.6 & 10.6 & 66.8 & 70.78 & 70.89 & 78.31 & 6.61 \\ 
    \hdashline
  MatCha \shortcite{liu2022matcha} & 90.20$^*$ & 38.20$^*$ & 64.20$^*$ & - & - & - & - & 12.20 & 39.40 & - & - & - & 64.00 & 60.90 & 85.21 & 83.40\\ 

  UniChart \shortcite{masry-etal-2023-unichart} & 88.56$^*$ & 43.92$^*$ & 66.24$^*$ & - & - & - & 14.88 & 12.48 & 38.21 & - & - &  - & - & - & 94.01 & 91.10\\ 

  T5 \shortcite{masry-etal-2022-chartqa, kantharaj2022opencqa} & - & - & 59.80$^*$ & - & - & - & 57.93 & - & - & - & - &  - & - & - & - & -\\
  
  VL-T5 \shortcite{masry-etal-2022-chartqa, kantharaj2022opencqa, 2023-vistext} & - & - & 59.12$^*$ & - & - & - & 59.80 & - & - & - & 32.90 &  - & - & - & - & - \\

  OCR-T5 \shortcite{kantharaj-etal-2022-chart, 2023-vistext} & - & - & - & - & - & - & - & 35.39 & - & - & 10.49 &  - & - & - & - & - \\

  ResNet + BERT \shortcite{akhtar-etal-2023-reading} & - & - & - & - & - & - & - & - & - & - & - & 62.70 & - & - & - & - \\

  ChartLLaMA \shortcite{han2023chartllama} & - & - & 69.66$^*$ & - & - & - & - & 40.71 & - & - & 14.23 & - & - & - & - & - \\

  ChartAssistant \shortcite{meng2024chartassisstant} & - & - & 79.90$^*$ & - & - & - & 15.50 & 41.00 & - & - & 15.20 & - & - & - & - & 92.00 \\

  Pix2struct \shortcite{lee2022pix2struct} & - & - & 56.05$^*$ & - & - & - & 12.70 & 38.00 & - & - & 10.30 & - & - & - & - & - \\

  ChartInstruct \shortcite{masry2024chartinstruct} & - & - & 72.00$^*$ & - & - & - & 16.71 & 43.53 & - & - & 13.83 & - & 72.65 & - & - & - \\

  ChartGemma \shortcite{masry2024chartgemma} & - & - & 80.16$^*$ & - & - & - & - & - & - & - & - & 70.33 & 72.17 & - & - & - \\
  
  \midrule
  
 \end{tabular}}
 \caption{An overview of the evaluation results on five tasks: ChartQA, Chart Summarization, OpenCQA, Chart-Fact-checking, and Chart-to-Table. Here, the ChartQA results with a `*' denote results without using CoT. The results except from Gemini, GPT-4V, Claude-3-haiku, and Phi-3-vision-inst, are noted based on the best-performing models as presented in the respective research paper.
 }
 \label{tab:overview-results}
\end{table*}

\begin{figure}[t]
    \centering
    \includegraphics[scale = 0.30]{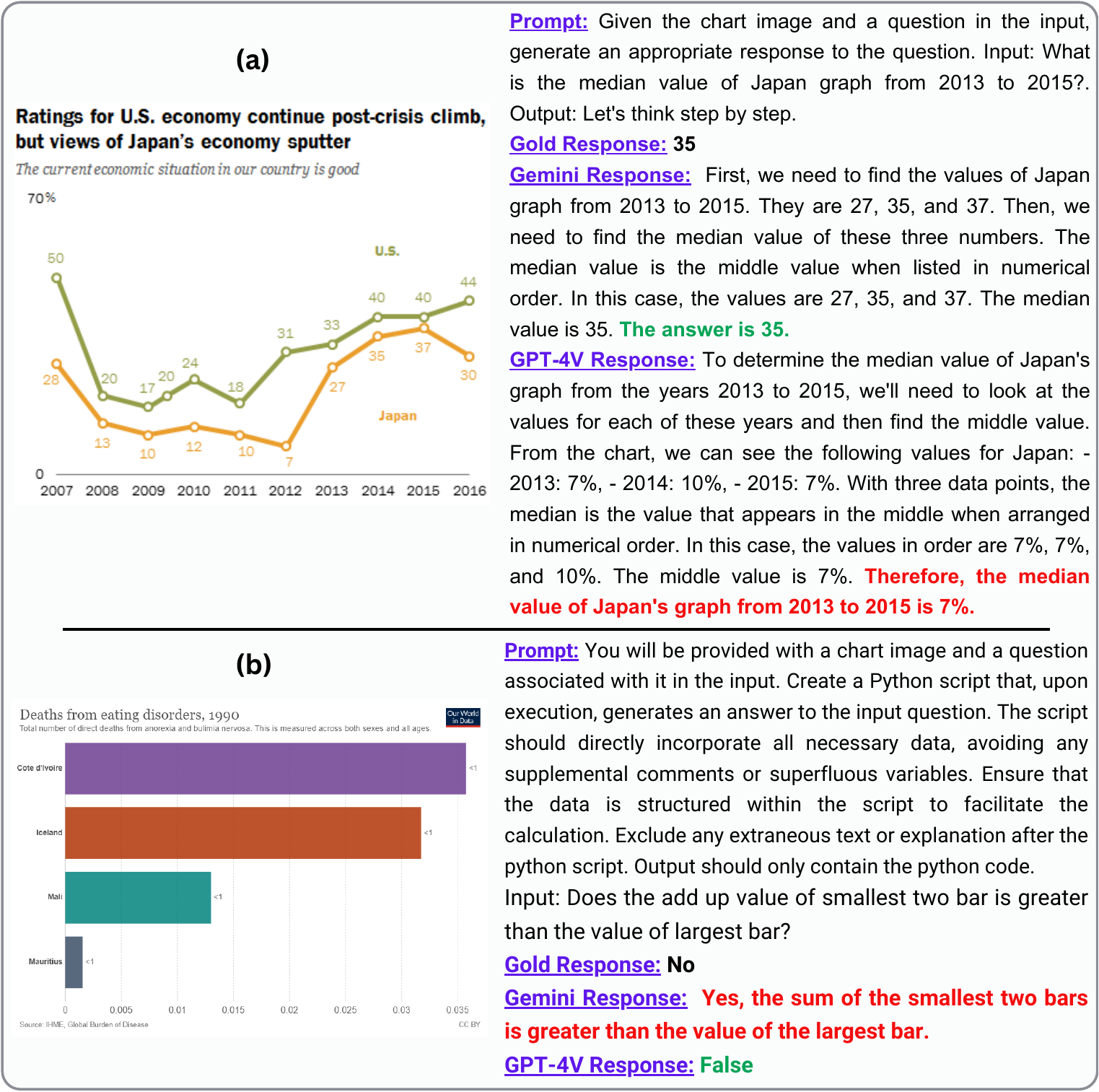}
    \caption{Figure (a) is where the Gemini is successful in 0-shot CoT, but the GPT-4V fails. Figure (b) shows the GPT-4V's success in PAL setup, while the Gemini fails. 
    Here, \textcolor{red}{\textbf{Red}} text denotes incorrect, and \textcolor{darkpastelgreen}{\textbf{Green}} text is correct.}
    \label{fig:cot-pal-gemr-gptr}
\end{figure}

\subsection{Performance in ChartQA task}
\label{subsec-reschartqa} 
We perform a quantitative evaluation of the LVLMs in ChartQA in two different prompting setups, \textbf{(i)} zero-shot Chain-of-Thought (CoT)\footnote{We report the zero-shot result of Phi-3 in \Cref{tab:overview-results} from the technical report \cite{abdin2024phi3} of the model.} \cite{wei2023chainofthought} prompting and \textbf{(ii)} prompting strategy introduced in Program-aided Language models (PAL) \cite{gao2023pal}. 
Initially, we evaluated the models' performance using the \emph{`relaxed accuracy'} metric as discussed in \cite{masry-etal-2023-unichart}. However, given the open-ended nature of the CoT responses, we conducted a manual evaluation to determine the actual accuracy of the models in the CoT setup. Also, we chose the PAL setup to examine whether separating the computation of complex queries and delegating them to a Python interpreter improves the performance of factoid question answering with charts.  
For further details on the datasets, refer to the \S\ref{app:dataset}. Below, we present our key findings: 

\paragraph{Performance in Zero-shot CoT.}\, In the case of zero-shot CoT, the Gemini outperformed GPT-4V by a margin of \num{3}\%
(Table \ref{tab:overview-results}), while Claude-3 performed the worst, achieving an average accuracy of \num{44.56}\%. 
\Cref{fig:cot-pal-gemr-gptr}(a) depicts an example case where a line chart about the economic situation of two countries is given and the models are asked: \textit{What is the median value of Japan graph from 2013 to 2015?}. With CoT reasoning, Gemini answered correctly, whereas GPT-4V answered incorrectly.

\paragraph{Performance in Program-Aided Reasoning.}\, In this setup, the LVLMs were tasked with generating Python code to answer questions based on specific charts. \Cref{tab:overview-results} demonstrates that GPT-4V and Claude-3 achieved relatively higher performance levels compared to Gemini, indicating their greater proficiency in consistently producing more effective and functional code.  
Gemini's lower accuracy is mostly due to its inability to generate executable code on an average of \num{35}\% of cases, across both ChartQA test sets. \Cref{fig:cot-pal-gemr-gptr}(b) depicts an example where a bar chart illustrates the deaths from eating disorders in 1990 in four different countries
and the models are prompted to answer the following question: \textit{Does the add up value of smallest two bars is greater than the value of the largest bar?} Using the PAL method, GPT-4V answered correctly, while Gemini answered incorrectly.

\paragraph{Dependence on Data labels.}\, For this experiment, we chose the two best performers in the ChartQA task in zero-shot CoT setup (\Cref{tab:overview-results}). As demonstrated in \Cref{chartqavis-automatic}, the absence of text labels that show data values diminishes the performance of both models, with GPT-4V being more affected.  
Moreover, GPT-4V frequently declines to respond when data labels are absent, as depicted in the right example in Figure \ref{fig:chartqavis_examples}. 
Our manual analysis suggests that these models exhibit better performance when the values of chart objects (e.g., bars, lines) align closely with the y-axis labels, leveraging these labels as a reference point, as illustrated in the left example in \Cref{fig:chartqavis_examples}.  
Conversely, a disparity between the visual element values and y-axis labels leads to poorer performance. These findings underscore a critical limitation in the capabilities of both Gemini and GPT-4V in interpolating the data values of the chart visual elements (e.g., bars, lines, pie) based on their visual attributes (e.g., heights, areas).

\subsection{Performance in Chart Summarization}
\label{subsec-rescharttotext}
We assess the text generation capabilities of 
LVLMs using both automatic metrics (see \Cref{tab:chart-to-text-automatic}, \Cref{tab:vistext-automatic}) and qualitative\footnote{Since most closed-source models do not support fine-tuning, we specifically conduct human evaluation only on closed-source models to check how they perform in zero-shot.} metrics.

\paragraph{Replication of Gold summaries.}\,
On the BLEU measure, Claude-3 and Gemini surpassed GPT-4V and Phi-3 in generating chart summaries that closely resemble the gold standard. However, in terms of BERTScore, all models performed similarly, suggesting 
identical performance when contextual similarity is considered instead of tokens.

\paragraph{Evidence of Factual errors in summaries.}\, Our qualitative evaluation of the three closed-source models across \num{100} samples suggests that all models produce fluent and coherent text. However, all models demonstrated factual errors. More specifically,  
GPT-4 exhibited the lowest overall error percentage (\num{5}\%), followed by Claude-3 (\num{16}\%), while Gemini had the highest error percentage (\num{27}\%).
Examples of factually incorrect summaries generated by these models are provided in \Cref{tab:app-c2t-pew} and \ref{tab:app-c2t-statista}.

\begin{figure*}[t]
    \centering
    \includegraphics[width=\textwidth]{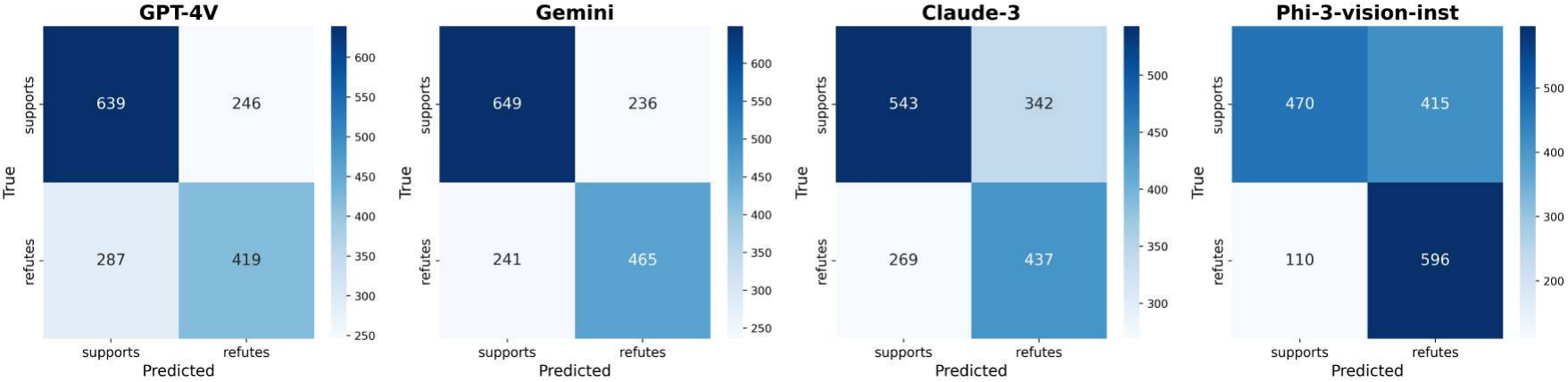}
\caption{Confusion matrices for different LVLMs on the ChartFC dataset.}
    \label{fig:conf_mat}
\end{figure*}

\paragraph{Inclusion of visual references in summaries.} \,
Referencing visual features in a chart can aid readers in coordinating between the chart and the text \cite{kim2020answering}. In contrast to Gemini (\num{25} samples out of \num{200}) and Claude-3 (\num{50} samples out of \num{200}), GPT-4V consistently references (\num{189} samples out of \num{200}) visual features of charts in its summaries, often using phrases such as `horizontal bars represent \dots'. 
Additionally, summaries generated by GPT-4V often contain incorrect references to color features (see \Cref{tab:app-c2t-visual-incorrect}). 
This inaccuracy is further evidenced by the fact that \num{80} of GPT-4V's \num{189} sentences contained errors, whereas Gemini and Claude-3 only had \num{24} and \num{7} errors respectively.

\paragraph{Identification of trends in charts.} Identifying and explaining important trends is critical in chart analysis~\cite{lundgard2021accessible}. \Cref{tab:app-c2t-correct-trend} demonstrates an example where both models correctly captured simple trends. However, our investigation indicates that Gemini is more prone to missing trends, while GPT-4V and Claude-3 tend to capture trends more effectively (see \Cref{tab:app-c2t-missing-trend}). 

\subsection{Performance in OpenCQA task}
\label{subsec-resopencqa}
Similar to the Chart-to-Text task, OpenCQA favors Gemini on all automatic metrics, except perplexity (\Cref{opencqa-automatic}).  
Our qualitative evaluation across 100 samples shows that responses from all closed-source models were fluent and coherent but contained factual errors. 
Claude-3 had factual inaccuracies in about 4\% of cases, GPT-4V in 5\%, and Gemini in 17\%. 
We observed that LVLMs, particularly GPT-4V, often generate detailed answers that include information that is not present in the gold answers but available in the chart. 
In our evaluation,
we specifically checked whether the LVLM-generated responses contradicted any information mentioned in the gold answers.

\subsection{Performance in Fact-checking task}
\label{subsec-resfactcheck}
We evaluated GPT-4V, Gemini, and Claude-3, and open-source models such as Phi-3 in the Fact-checking with charts task in the ChartFC \cite{akhtar-etal-2023-reading}, and the ChartCheck dataset \cite{akhtar2023chartcheck} (see \Cref{tab:overview-results}).
We observed that all the evaluated models performed remarkably in the ChartCheck dataset, beating the existing SoTA Matcha \cite{liu2022matcha} by some margin (see \Cref{tab:overview-results}) in both of the test sets. Similarly, in the ChartFC dataset, GPT-4V, Gemini, and Phi-3 performed better than the existing SoTA reported in \cite{akhtar-etal-2023-reading}, except the Claude-3 model. Our investigation also revealed that open-source models like Phi-3 produced more false negatives compared to their closed-source counterparts in this dataset (see \Cref{fig:conf_mat}).
Nonetheless, in both benchmarks, the average accuracy is below 72\%, indicating potential for further improvement in this task.

\subsection{Performance in Chart-to-Table task}
\label{subsec-charttotable}
Chart-to-Table 
requires the model to extract the underlying data table from the provided chart image. To assess LVLMs capabilities in this task, we utilize the ChartQA dataset \cite{masry-etal-2022-chartqa} which provides the underlying data tables for the chart image. 
As depicted in \Cref{tab:overview-results}, notably, GPT4-V demonstrates superior performance in RMS, emphasizing its capability to accurately reconstruct the structure of tables derived from charts. Conversely, Gemini exhibits higher proficiency in RNSS, indicating its strength in accurately estimating numerical values from chart images.

\begin{table*}[t!]
\centering
\resizebox{\textwidth}{!}{%
\begin{tabular}{l|p{0.8\textwidth}|llllll}
\hline
\multirow{3}{*}{Error Type} &
  \multicolumn{1}{c|}{\multirow{3}{*}{Example}} &
  \multicolumn{6}{c}{Average Error Count (Per Summary)} \\ \cline{3-8} 
 &
  \multicolumn{1}{c|}{} &
  \multicolumn{3}{c|}{Pew} &
  \multicolumn{3}{c}{Statista} \\ \cline{3-8} 
 &
  \multicolumn{1}{c|}{} &
  \multicolumn{1}{l|}{Gemini} &
  \multicolumn{1}{l|}{GPT-4V} &
  \multicolumn{1}{l|}{Claude 3 Haiku} &
  \multicolumn{1}{l|}{Gemini} &
  \multicolumn{1}{l|}{GPT-4V} &
  \multicolumn{1}{l}{Claude 3 Haiku} \\ \hline \hline
Entity &
  Alberta is the top producer, with 126,082,558 \textcolor{red}{billion} cubic meters of natural gas. &
  \multicolumn{1}{c|}{\textbf{0.47}} &
  \multicolumn{1}{c|}{0.51} &
  \multicolumn{1}{c|}{1.39} &
  \multicolumn{1}{c|}{\textbf{0.66}} &
  \multicolumn{1}{c|}{0.88} &
  \multicolumn{1}{c}{1.85} \\ \hline
Relation &
  The population density was \textcolor{orange}{lowest} in 2018 and \textcolor{orange}{highest} in 1960 &
  \multicolumn{1}{c|}{\textbf{0.16}} &
  \multicolumn{1}{c|}{0.17} &
  \multicolumn{1}{c|}{0.17} &
  \multicolumn{1}{c|}{0.17} &
  \multicolumn{1}{c|}{0.21} &
  \multicolumn{1}{c}{\textbf{0.12}}  \\ \hline
Subjective &
  The chart shows that the number of cases is significantly higher in urban areas compared to rural areas. &
  \multicolumn{1}{c|}{0.02} &
  \multicolumn{1}{c|}{0.02} &
  \multicolumn{1}{c|}{\textbf{0.01}} &
  \multicolumn{1}{c|}{0.02} &
  \multicolumn{1}{c|}{0.02} &
  \multicolumn{1}{c}{\textbf{0.00}} \\ \hline
Contradictory &
  There is a clear \textcolor{green}{upward trend in the number of deaths caused by influenza and pneumonia over time. This trend is likely due to improvements in public health measures, such as vaccination and sanitation}. &
  \multicolumn{1}{c|}{0.19} &
  \multicolumn{1}{c|}{\textbf{0.12}} &
  \multicolumn{1}{c|}{0.15} &
  \multicolumn{1}{c|}{0.29} &
  \multicolumn{1}{c|}{\textbf{0.14}} &
  \multicolumn{1}{c}{0.19} \\ \hline
Unverifiable &
  Overall, the increase of percentage of people who have completed high school, has a positive impact on the United States. &
  \multicolumn{1}{c|}{\textbf{0.03}} &
  \multicolumn{1}{c|}{\textbf{0.03}} &
  \multicolumn{1}{c|}{\textbf{0.03}} &
  \multicolumn{1}{c|}{0.05} &
  \multicolumn{1}{c|}{0.04} &
  \multicolumn{1}{c}{\textbf{0.03}}  \\ \hline
Invented &
  The unemployment rate increased sharply from 3.3\% in November 2019 to 15.7\% in April 2020, \textcolor{gold}{the highest level since the Great Recession.} &
  \multicolumn{1}{c|}{\textbf{0.02}} &
  \multicolumn{1}{c|}{0.07} &
  \multicolumn{1}{c|}{0.03} &
  \multicolumn{1}{c|}{\textbf{0.03}} &
  \multicolumn{1}{c|}{0.05} &
  \multicolumn{1}{c}{0.04}  \\ \hline
Total &
   &
  \multicolumn{1}{c|}{\textbf{0.89}} &
  \multicolumn{1}{c|}{0.92} &
  \multicolumn{1}{c|}{1.76} &
  \multicolumn{1}{c|}{1.26} &
  \multicolumn{1}{c|}{1.35} &
  \multicolumn{1}{c}{2.23}  \\ \hline
\end{tabular}%
}
\caption{Color-coded table example of hallucinations detected in chart summaries by FAVA.  Key: \textcolor{red}{Red} = entity hallucination; \textcolor{orange}{Orange} = relation hallucination; \textcolor{green}{Green} = contradictory hallucination; \textcolor{gold}{Gold} = invented hallucination. Subjective and unverifiable hallucinations exist at the sentence level and are not highlighted. Average error counts per type are included.}
\label{tab:fava-c2t-example}
\end{table*}

\subsection{Hallucination Analysis}
\label{subsec-c2t-hallucination-analysis} 
\change{To analyze hallucinations in LLM-generated responses, we sampled the chart summaries generated by Gemini, GPT-4V, and Claude-3 in the Chart-to-Text data. We used the Factuality-Aware Visual Analytics (FaVA) \cite{mishra2024finegrained} methodology for hallucination detection, by categorizing hallucinations into entity, relation, subjective, contradictory, unverifiable, and invented types (see a color-coded example in \Cref{tab:fava-c2t-example}). 
The analysis showed that the \texttt{entity} category had the highest error count among all categories, which is consistent with findings in other NLP tasks~\cite{mishra2024finegrained}. Substantial errors also come from \texttt{Relation} and \texttt{contradictory} categories. 
Overall, Claude-3 had the highest total error count (\num{1.76} for Pew, \num{2.23} for Statista), while Gemini (\num{0.89} for Pew, \num{1.26} for Statista) and GPT-4V (\num{0.92} for Pew, \num{1.35} for Statista) had fewer errors.
The above finding highlights the urgent need to study and detect the frequent types of hallucinations (\texttt{entity} and \texttt{relations}) which are often phrase-level and
can be fixed by minimal editing erroneous phrases~\cite{chen2023purr}.
}

\begin{table}[t]
\centering
\caption{The performance of GPT-4V and Gemini in answering questions (Accuracy) and generating sentences across various semantic levels. `Coverage' indicates average sentences per semantic level in summaries.}
\label{tab:sem_content}
\resizebox{\columnwidth}{!}{%
\begin{tabular}{lcc|cc}
\textbf{} & \multicolumn{2}{c}{\textbf{Coverage}} & \multicolumn{2}{c}{\textbf{Accuracy (\%)}} \\\hline
\textbf{Semantic Level}                              & GPT-4V & Gemini & GPT-4V & Gemini \\\midrule
\textit{\textbf{L1}: Visual encodings}               & \textbf{1.69}   & 1.25   & \textbf{70.0}   & 57.5   \\
\textit{\textbf{L2}: Statistical and relational}     & 0.56   & \textbf{0.87}   & \textbf{80.5}   & 62.0   \\
\textit{\textbf{L3}: Perceptual and cognitive}       & \textbf{0.70}   & 0.41   & \textbf{58.9}   & 48.2   \\
\textit{\textbf{L4}: contextual and domain-specific} & 0      & \textbf{0.03}   & 15.5   & \textbf{16.0}  \\\Xhline{1pt}
\end{tabular}%
}
\end{table}
\subsection{Analysis of Semantic Levels}
\label{subsec-semantics}
For text generation tasks (e.g., chart summarization), a crucial question is how different semantic contents are covered in output texts and how accurately models can understand such statements.  
We analyze this question using the four-level semantic framework~\cite{lundgard2021accessible} as explained in \Cref{subsec-eval}. Research suggests that readers prefer chart summaries that describe more high-level trends, patterns, and contextual explanations (Levels 3 \& 4) over low-level information, e.g., chart type, axes, color encodings, and simple statistics like averages and extrema (Levels 1 \& 2)~\cite{stokes2022striking}. However, low-level information might be useful for some chart accessibility applications.

\begin{figure}[t]
    \centering
    \includegraphics[width=\columnwidth]{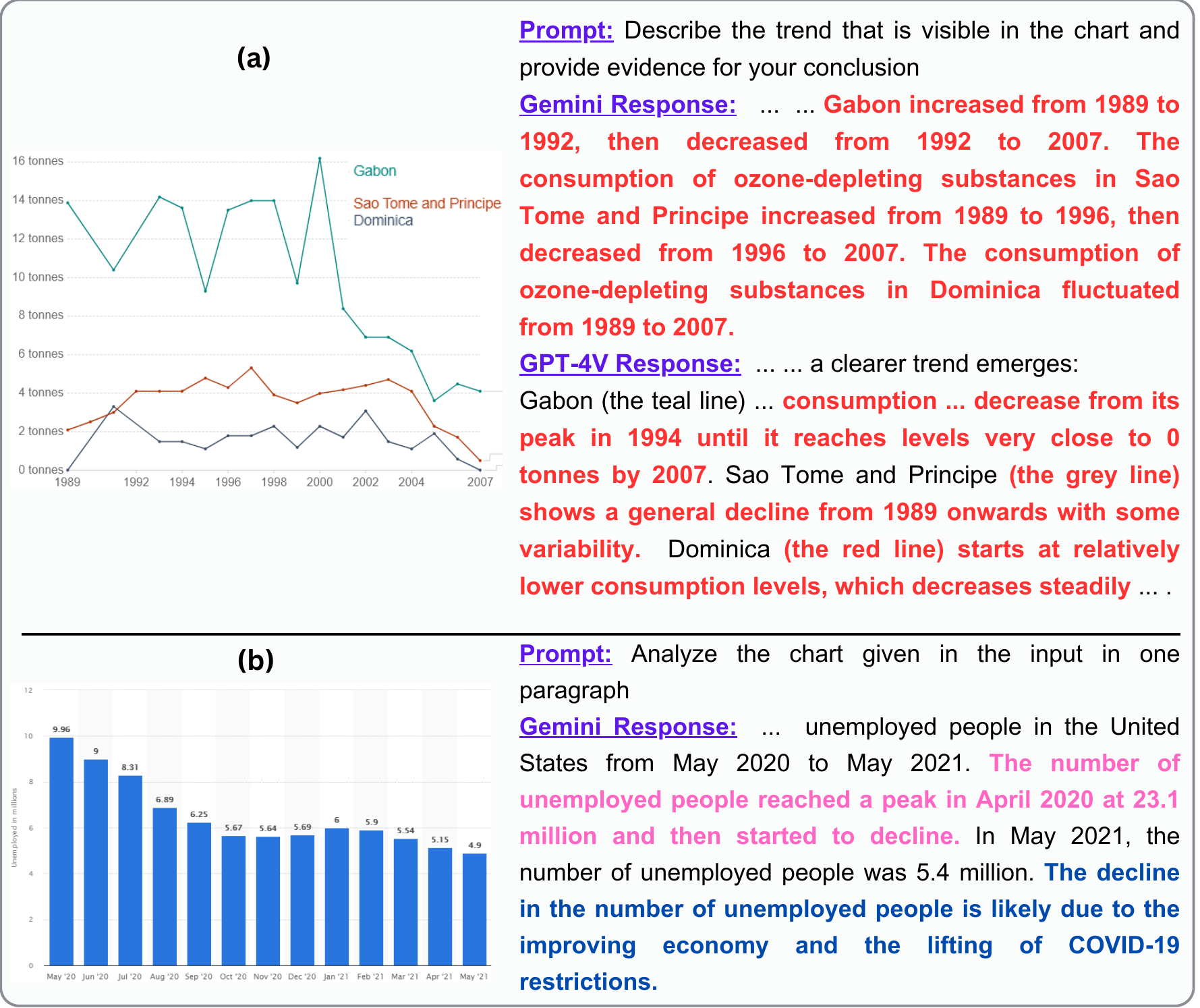}
\caption{Both Gemini and GPT-4V fail to identify trends (a). Out-of-context but relevant information generated by Gemini (b). Here, \textcolor{red}{\textbf{Red}} text indicates incorrect facts, \textcolor{fluorescentpink}{\textbf{Pink}} text denotes out-of-context, and \textcolor{blue}{\textbf{Blue}} text represents domain-specific details. \textit{`\dots'} indicates abbreviated text for brevity.}
    \label{fig:sem_errors}
\end{figure}

\paragraph{Generating different semantic contents.}\, We manually examine model-generated texts for \num{200} chart-to-text samples to understand how they cover different types of semantic content.  From \Cref{tab:sem_content} and \Cref{fig:c2t-coverage-sem}, we observe that  GPT-4V produces longer summaries of chart-specific visual information (Levels 1 \& 3) while  Gemini produces concise summaries with some statistical and domain-specific information (Levels 2 \& 4) and Claude's outputs fall in-between these two models (more details in \S\ref{app:c2t-four-level-sem-analysis}). 
We also observe that GPT-4V not only produces statements describing high-level trends but also does so with higher accuracy than other models  (see error examples in \Cref{fig:sem_errors}).  Another important observation is that all models fail to include sufficient contextual and domain-specific information (Level 4) that explains trends and patterns in charts using external domain information (e.g., social and political contexts), which human authors often include in high-quality chart descriptions (e.g., Pew chart summaries).

\paragraph{Understanding different semantic contents.}\,  In another experiment, we examine LVLMs' ability to understand and answer questions across different types of semantics. To this end, we created \num{200} different question prompts for each of the four semantic levels using charts from the ChartQA dataset. We chose Gemini and GPT-4V as they are the top-performing closed-source models (see experimental details in \S\ref{app:4-level-sem}).

From \Cref{tab:sem_content}, we observe that GPT-4V outperforms Gemini in answering questions across all levels except for Level 4, in which their performance is similar. Both models struggle to describe complex trends in line charts with multiple, highly fluctuating lines. \Cref{fig:sem_errors}(a) illustrates such a scenario, where the chart indicates that \textit{Ozone-depleting substance consumption in Gabon peaked in 2000}, but both GPT-4V and Gemini suggest otherwise.

Another interesting observation is that Gemini can
extrapolation of factually accurate insights beyond the chart data. For example, in \Cref{fig:sem_errors}(b), although the x-axis labels of the bars began in May 2020, Gemini managed to describe trends by including previous years by outputting \textit{``...The number of unemployed people reached a peak in April 2020 at 23.1 million and then started to decline.''} While this information was not directly evident in the chart data, it aligns closely with statistics from the U.S. Bureau of Labor Statistics \cite{uslabor}. This finding is consistent with the observation that Gemini can cover more contextual and domain information from external sources.

\begin{figure}[t]
    \centering
    \includegraphics[scale = 0.37]{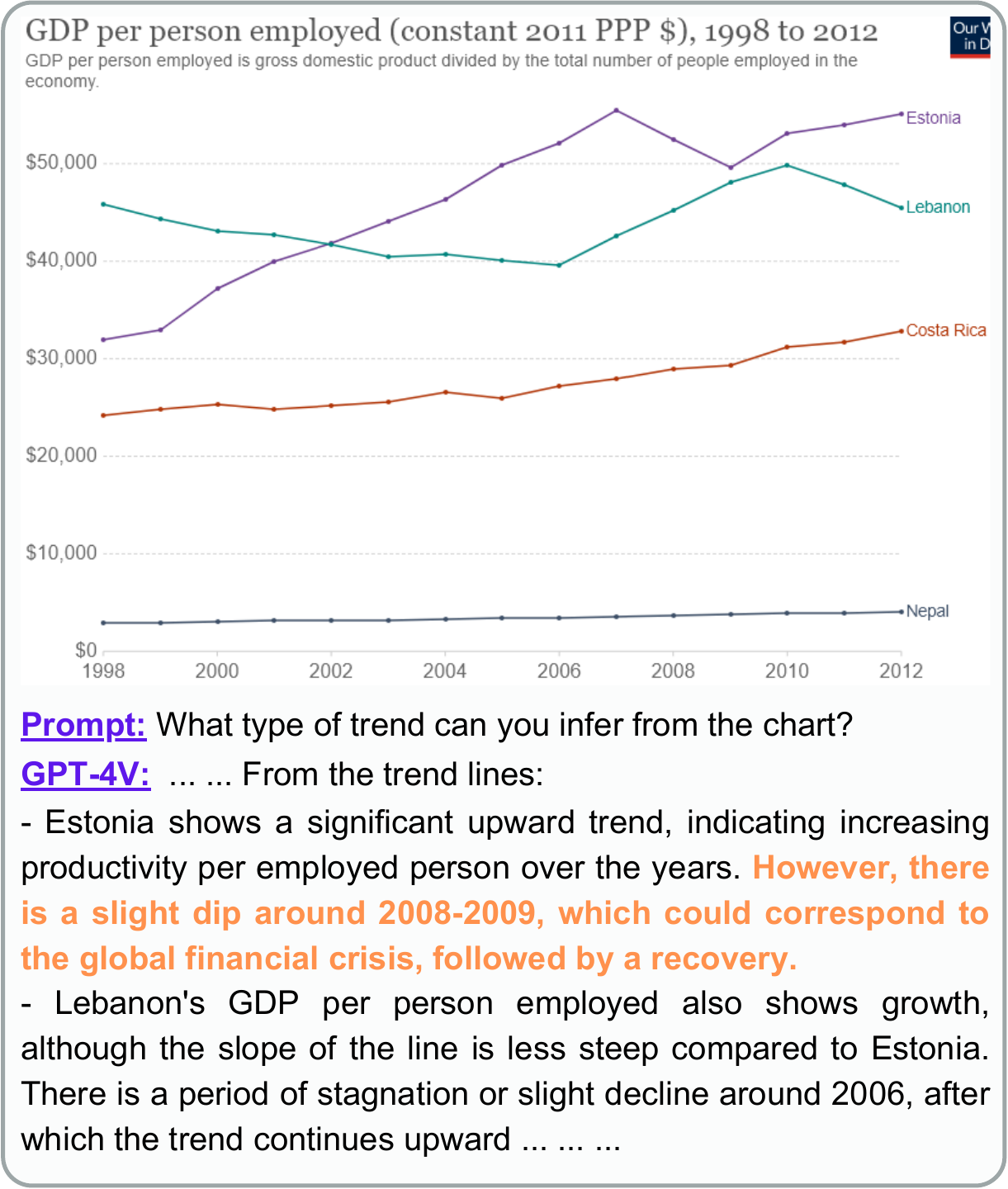}
    \caption{The figure is where the GPT-4V model shows some potential bias in the response. Here, \textcolor{orange}{\textbf{Orange}} text denotes potentially biased output that is out-of-context of the chart data.\textit{`\dots'} indicates abbreviated text for brevity.}
    \label{fig:modelbias}
\end{figure}

\subsection{Potential bias in Model responses} 
\label{sec:bias}
A notable concern with LVLMs is their potential for biased output generation~\cite{openai2023gpt4}. However, this critical issue remains unexplored in the chart domain. Prior research such as \cite{huang2024pixels} has mentioned some issues of LVLMs, including coverage of chart content, fairness, and bias, but did not provide conclusive evidence directly confirming these issues. To bridge this gap, in a preliminary experiment, we manually analyzed 200 responses from GPT-4V, the top performer in understanding high-level semantics. We found several cases where the model's causal explanations might reflect training data biases. For example, in \Cref{fig:modelbias}, the model attributed Estonia's slight GDP dip in 2008-2009 to the global financial crisis, which could be a spurious correlation. While Estonia's data did show a slight dip around that time, attributing it solely to the global financial crisis might be inaccurate since Lebanon and Costa Rica also experienced GDP increases during this period. 

This finding highlights a pressing need to deeply explore the bias problem in the chart domain. An initial solution could be implementing measures like \textit{pre-processing} (altering model inputs), \textit{in-training} (adjusting the optimization process), \textit{intra-processing} (changing inference behavior), and \textit{post-processing} (rewriting model outputs)~\cite{gallegos2024bias}.

\section{Conclusion and Future Directions}
This study presents a comprehensive analysis of LVLMs (GPT-4V, Gemini, Claude, and Phi-3) in interpreting and deriving insights from chart images in real-world scenarios, where data tables may not be available.  Through qualitative and quantitative analyses, we evaluate these models across various tasks, including zero-shot CoT prompting and program-aided reasoning, assessing their impact on chart question-answering tasks. Additionally, we examine LVLMs' performance in open-ended text generation from chart-related tasks.

These analyses highlight both the strengths and limitations of LVLMs and identify key research gaps. First, enhancing the generalizability and reasoning abilities of open-source LVLMs in chart-related tasks is a priority that can be explored via instruction tuning~\cite{masry2024chartinstruct}. Second, there is significant potential for LVLMs to produce semantically rich texts that describe high-level trends and contextual information more effectively. Third, addressing key issues such as hallucinations, factual errors, and bias requires developing new benchmarks and models for detection and mitigation. We hope that the insights gained from this study will catalyze further research and advancements in the emerging area of chart reasoning.

\section*{Limitations}
Since the pretraining corpus of both the large vision language models (LVLMs) is unknown (not open-source), some of the datasets used for evaluation may or may not appear in the pretraining data or instruction tuning data of the models. Although we covered all the important tasks, i.e., Chart Summarization, Chart Question-Answering, Open-ended Chart Question-Answering, and Fact Checking with Charts, etc., there are some tasks, i.e., Chart-to-table not addressed in this research. At the time of evaluation, we did not provide any underlying data table corresponding to the chart in the input. However, our motivation for this research was to show how different state-of-the-art LVLMs perform when the underlying data table is not present for chart understanding tasks, which is often the case in real-world scenarios. Further, variations of charts and labels are limited due to the open-sourced datasets available for the tasks. We did not perform the qualitative evaluation in the ChartQA task, since the task is based on factoid-QA about Charts and only requires single token answers (either text or a numerical value), for which automatic evaluation is sufficient.

\section*{Ethics Statement}
This study independently evaluated LVLMs' responses without involving any external parties, hence, no extra financial compensation was necessary. The authors themselves performed all the human assessments presented in this paper. As the focus of the research was solely on assessing LVLM's capabilities, effectiveness, and limitations in several chart understanding tasks, the human evaluation performed by the authors does not add any ethical issues or unwanted biases. Further, the datasets utilized in this study are all open-sourced academic datasets, thus licensing was not required. Additionally, no information has been used that can directly relate to the identification of any person while evaluating the responses from LVLMs.

\section*{Acknowledgement}
The authors would like to thank the anonymous reviewers for their helpful comments and the meta-reviewer for constructive suggestions in improving this work. This research was supported by the Natural Sciences and Engineering Research Council (NSERC), Canada, Canada Foundation for Innovation, and the CIRC  grant on Inclusive and Accessible Data Visualizations and Analytics.  
We also thank Compute Canada for providing us with the computing resources to conduct experiments, as well as Anthropic for providing us free access to the
Claude-3 API. 

\bibliographystyle{acl_natbib}
\bibliography{chart2text}
\newpage
\appendix
\section{Appendices}
\subsection{Datasets}
\label{app:dataset}

\subsubsection{ChartQA}
In our study, we employ the test set from the ChartQA dataset, as introduced by \citet{masry-etal-2022-chartqa}. The test set of the dataset is composed of two primary categories of questions: those created by humans and those augmented by models. Specifically, the set of human-generated questions includes 625 distinct charts with 1250 corresponding question-answer pairs. Similarly, the model-generated, or augmented set, comprises 987 unique charts and 1250 question-answer pairs.

\subsubsection{ChartQA$^*$}
\label{app:chartqa-no-label}
\begin{figure*}[t]
    \centering
    \includegraphics[width=\columnwidth]{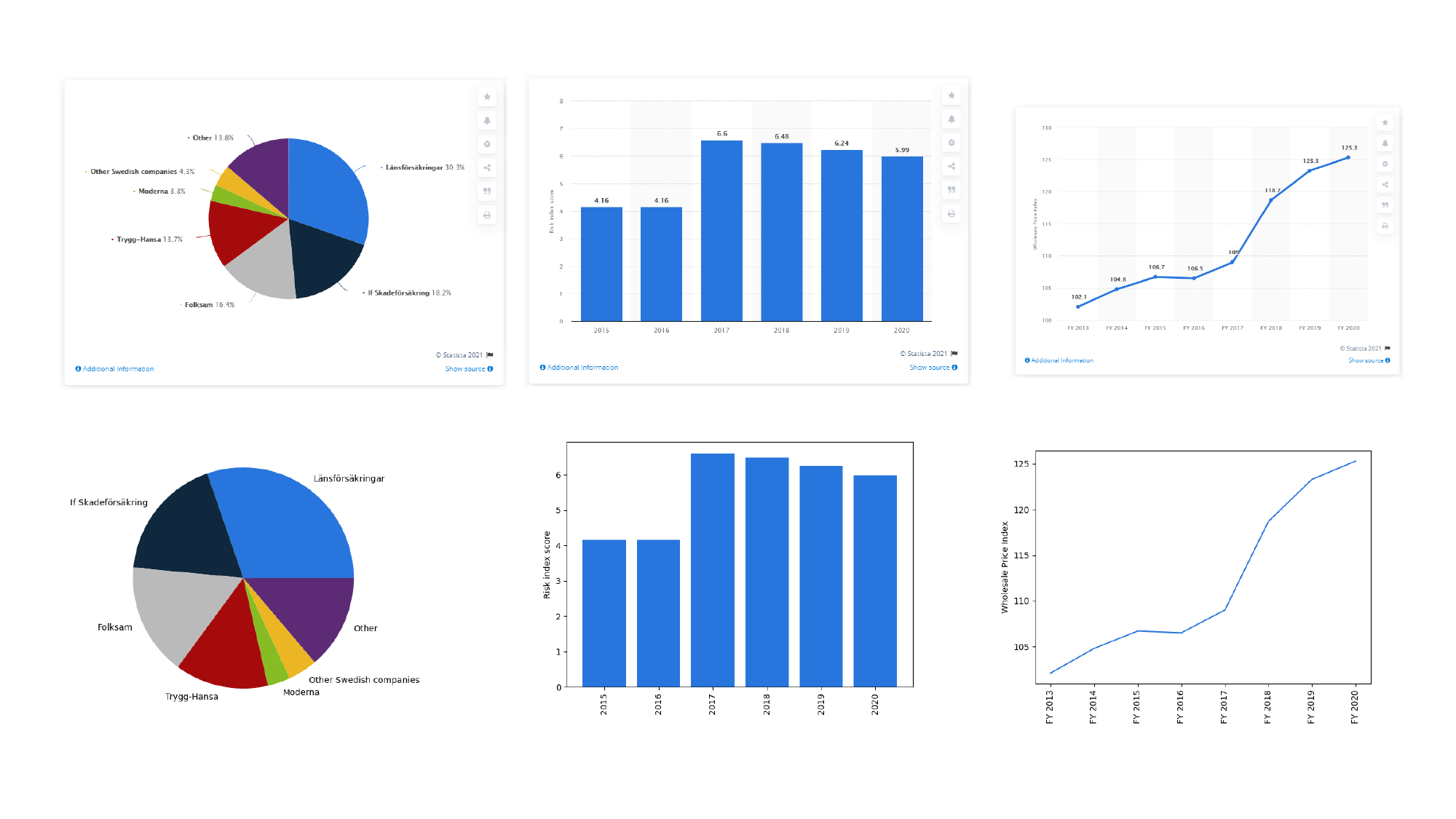}
    \caption{Examples of charts with and without the data labels.}
    \label{fig:chartqavisbeforeafter}
    \vspace{-3mm}
\end{figure*}

We introduce this dataset as a variation of the ChartQA dataset, in which charts do not explicitly show data values as labels near the corresponding chart elements (e.g., bars, lines), rather the model needs to estimate these values from the chart (e.g., based on bar heights and axis labels). We introduce this setup to see whether LVLMs' performance relies on the explicit labels of the data values rather than their ability to recover data values from the visual elements in the chart. For this purpose, we modified the ChartQA dataset using Matplotlib~\cite{Hunter:2007}, removing the data labels from the chart images while keeping everything else the same (see examples in Figure \ref{fig:chartqavisbeforeafter}). Of the 1509 chart images in the test set, 1340 were successfully redesigned. The remaining 169 images were excluded due to missing metadata. 

\subsubsection{ChartFC}
For the fact-checking with charts task, we utilize the ChartFC dataset from \citet{akhtar-etal-2023-reading}. The dataset is structured so that each entry contains a claim phrased in natural language, a related chart image, and a label that falls into one of two categories: \textit{`supports'} or \textit{`refutes'}. We evaluate the LVLMs in the test set of the dataset, which contains 885 examples belonging to the \textit{`supports'} class and 706 examples belonging to the \textit{`refutes'} class. The dataset comprises bar charts with different types, i.e., horizontal/vertical.

\subsubsection{ChartCheck}
For the fact-checking with charts task, we utilize another dataset from \citet{akhtar2023chartcheck}. The dataset is structured so that each entry contains a claim phrased in natural language, a related chart image, and a label that falls into one of two categories: \textit{`True'} or \textit{`False'}. The dataset contains two test sets, where test set-1 contains 937 samples and test set-2 contains 981 samples. We evaluated the models in both of these test samples.

\subsubsection{Chart-to-Text}
In our chart summarization study, we utilize the Chart-to-Text \cite{kantharaj-etal-2022-chart} benchmark. This benchmark encompasses two primary datasets: Statista\footnote{https://www.statista.com/} and Pew\footnote{https://www.pewresearch.org/}. Each sample within both datasets comprises a chart image, an underlying data table, a chart title, and a human-written gold summary. For our experimental purposes, we utilize the complete test split from each dataset, encompassing 1,393 samples from Pew and 5,222 samples from Statista.

\subsubsection{Vistext}
The VisText dataset \cite{2023-vistext} includes 12,441 pairs of charts and descriptive captions. The data tables in the dataset were sourced from the Statista corpus of the Chart-to-Text \cite{kantharaj-etal-2022-chart} benchmark. The dataset includes two different types of captions, i.e., L1 and L2/L3 which correspond to the semantic levels of \cite{lundgard2021accessible}. These captions offer insights into the charts' construction, highlight important statistics, and point out perceptual and cognitive phenomena. Each chart in VisText is represented in three ways: as a rasterized image, as a data table, and as a scene graph, which is a structured representation of the chart's visual elements akin to the Document Object Model (DOM) used in web pages.

\subsubsection{OpenCQA}
To study LVLMs performance on the Open-ended Chart Question-Answering task, we utilize the benchmark dataset OpenCQA from Kantharaj et al. \cite{kantharaj2022opencqa}. The dataset contains five different types of charts, i.e., bar, line, area, scatter, and pie. For our experiments, we use the test set from the dataset which comprises 1159 charts and 1159 question-answer pairs.

\begin{table}[h]
\centering
\resizebox{\textwidth}{!}{%
\begin{tabular}{l|cccc}
\hline
  \textbf{Model} &
  \textbf{BLEURT} (↑) &
  \textbf{CIDEr} (↑) &
  \textbf{PPL} (↓) &
  \textbf{BERTScore} (↑) \\ \hline \hline
Gemini  & \textbf{-0.28} & \textbf{1.88} & 2.06 & \textbf{0.87} \\ \hline
GPT-4V  & -0.45 & 1.63 & \textbf{1.85} & 0.85 \\ \hline
\end{tabular}%
}
\caption{ Evaluation results for different models on  OpenCQA . ↑ : Higher is better, ↓ : Lower is better.}
\label{opencqa-automatic}
\end{table}

\begin{table}[h]
\small
\centering
{%
\begin{tabular}{l|cc}
\hline
  \textbf{Model} &
  \textbf{ChartQA} &
  \textbf{ChartQA$^*$} \\ \hline \hline
Gemini   & \textbf{52.04} & 38.53 ({\color[HTML]{CB0000} $\downarrow$ 13.51\%}) \\ \hline
GPT-4V & \textbf{57.51} & 20.52  ({\color[HTML]{CB0000} $\downarrow$ 36.99\%}) \\ \hline
\end{tabular}%
}
\caption{ Relaxed Accuracy (RA) different models on the \textbf{ChartQA}$^*$ vs \textbf{ChartQA} test set. Here, ChartQA$^*$ denotes the charts from the test set of the ChartQA dataset without the annotations. Drop in performance compared to \textbf{ChartQA} is presented in round brackets.}
\label{chartqavis-automatic}
\end{table}

\begin{table}[h]
\centering
\resizebox{\textwidth}{!}{%
\begin{tabular}{l|cccccccc}
\hline
\multirow{2}{*}{\textbf{Model}} &
  \multicolumn{2}{c}{\textbf{BLEURT} (↑)} &
  \multicolumn{2}{c}{\textbf{CIDEr} (↑)} &
  \multicolumn{2}{c}{\textbf{PPL} (↓)} &
  \multicolumn{2}{c}{\textbf{BERTScore} (↑)} \\ \cline{2-9} 
          & Pew           & Stat & Pew  & Stat & Pew  & Stat & Pew  & Stat \\ \hline \hline
Gemini   & \textbf{-0.30} & -0.30 & \textbf{1.79} & 1.90  & 1.61 & 1.70  & \textbf{0.87} & 0.86 \\ 
GPT-4V & \textbf{-0.30} & -0.40 & 1.34 & 1.28 & 1.69 & 1.75 & 0.85 & 0.85 \\ 
Claude-3-Haiku   & -0.31 & \textbf{-0.25} & 1.56 & \textbf{1.91}  & 1.72 & 1.75  & \textbf{0.87} & \textbf{0.89} \\ 
\hline
  Phi-3-vision-128k-instruct & -0.88 & -0.49 & 1.47 & 1.54  & \textbf{1.49} & \textbf{1.51}  & 0.85 & 0.86 \\ 
\hline
\end{tabular}%
}
\caption{Detailed automatic evaluation results for different models on the Chart-to-Text dataset for Chart Summarization. ↑ : Higher is better, ↓ : Lower is better.}
\label{tab:chart-to-text-automatic}
\end{table}

\begin{table}[h]
\centering
\resizebox{\textwidth}{!}{%
\begin{tabular}{l|cccccccc}
\hline
\multirow{2}{*}{\textbf{Model}} &
  \multicolumn{2}{c}{\textbf{BLEURT} (↑)} &
  \multicolumn{2}{c}{\textbf{CIDEr} (↑)} &
  \multicolumn{2}{c}{\textbf{PPL} (↓)} &
  \multicolumn{2}{c}{\textbf{BERTScore} (↑)} \\ \cline{2-9} 
          & Pew           & Stat & Pew  & Stat & Pew  & Stat & Pew  & Stat \\ \hline \hline
Gemini   & -0.25 & -0.99 & 2.62 & \textbf{1.17}  & 1.83 & 1.82  & \textbf{0.88} & \textbf{0.87}\\ 
GPT-4V & -0.11 & -0.98 & 2.02 & 0.99 & 1.77 & 1.94 & 0.87 & 0.86 \\ 
Claude-3-Haiku   & -0.16 & -0.97 & 2.51 & \textbf{1.13}  & 1.85 & 1.85  & \textbf{0.88} & \textbf{0.87} \\ 
\hline
Phi-3-vision-128k-instruct & \textbf{-0.09} & -1.19 & 2.96 & 1.13  & \textbf{1.48} & \textbf{1.49}  & 0.88 & 0.85 \\ 
\hline
\end{tabular}%
}
\caption{Detailed automatic evaluation results for different models on the Vistext dataset for Chart Summarization. ↑ : Higher is better, ↓ : Lower is better.}
\label{tab:vistext-automatic}
\end{table}

\begin{figure*}[h]
    \centering
    \includegraphics[scale = 0.45]{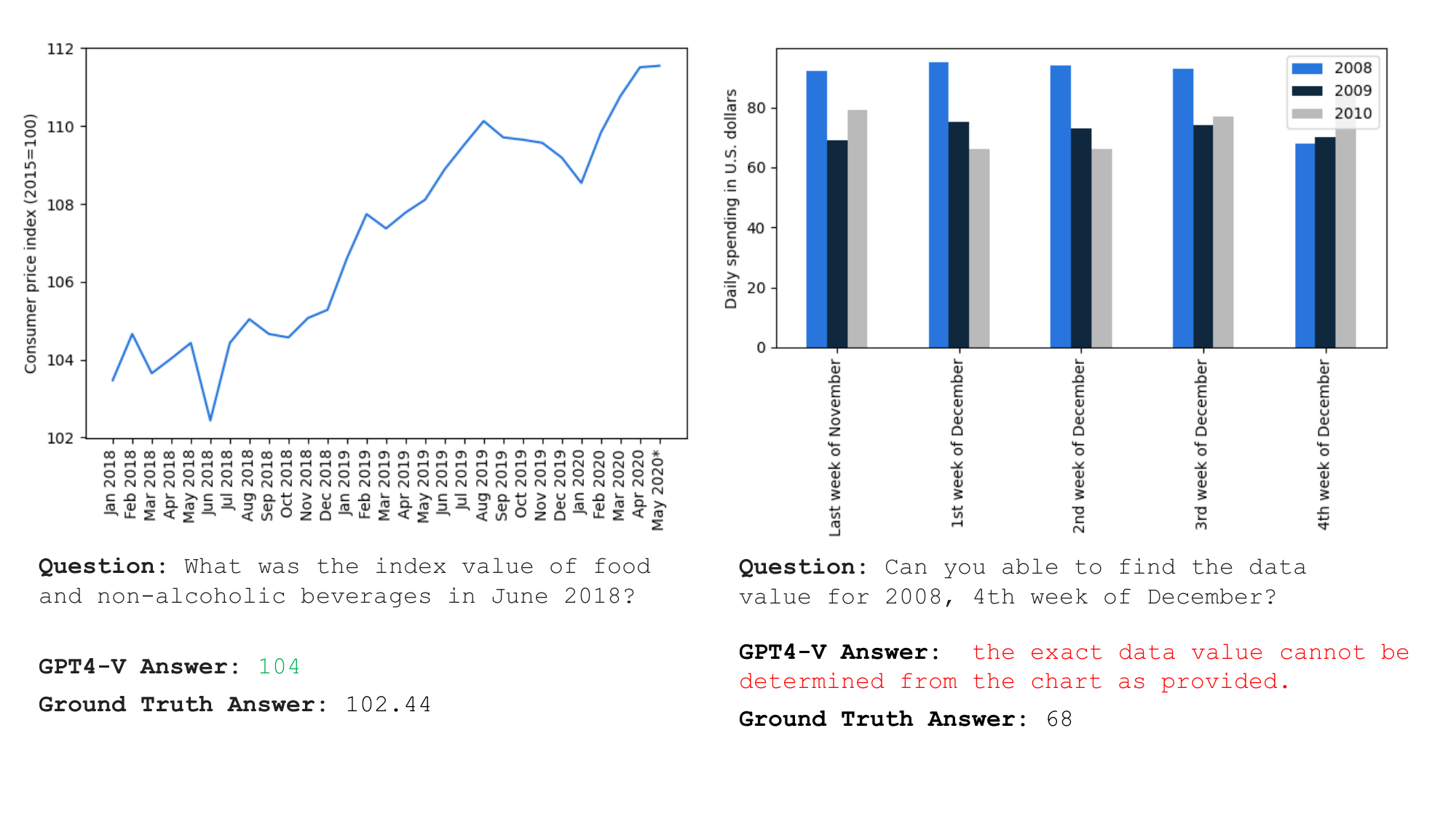}
    \caption{Sample outputs from GPT-4V on the ChartQA* benchmark.}
    \label{fig:chartqavis_examples}
\end{figure*}
\subsection{Analysis of 4-level Semantics}
\subsubsection{Coverage of 4-level semantic contents}
\label{app:c2t-four-level-sem-analysis}

\begin{figure*}[t!]
    \centering
    \includegraphics[width=\textwidth]{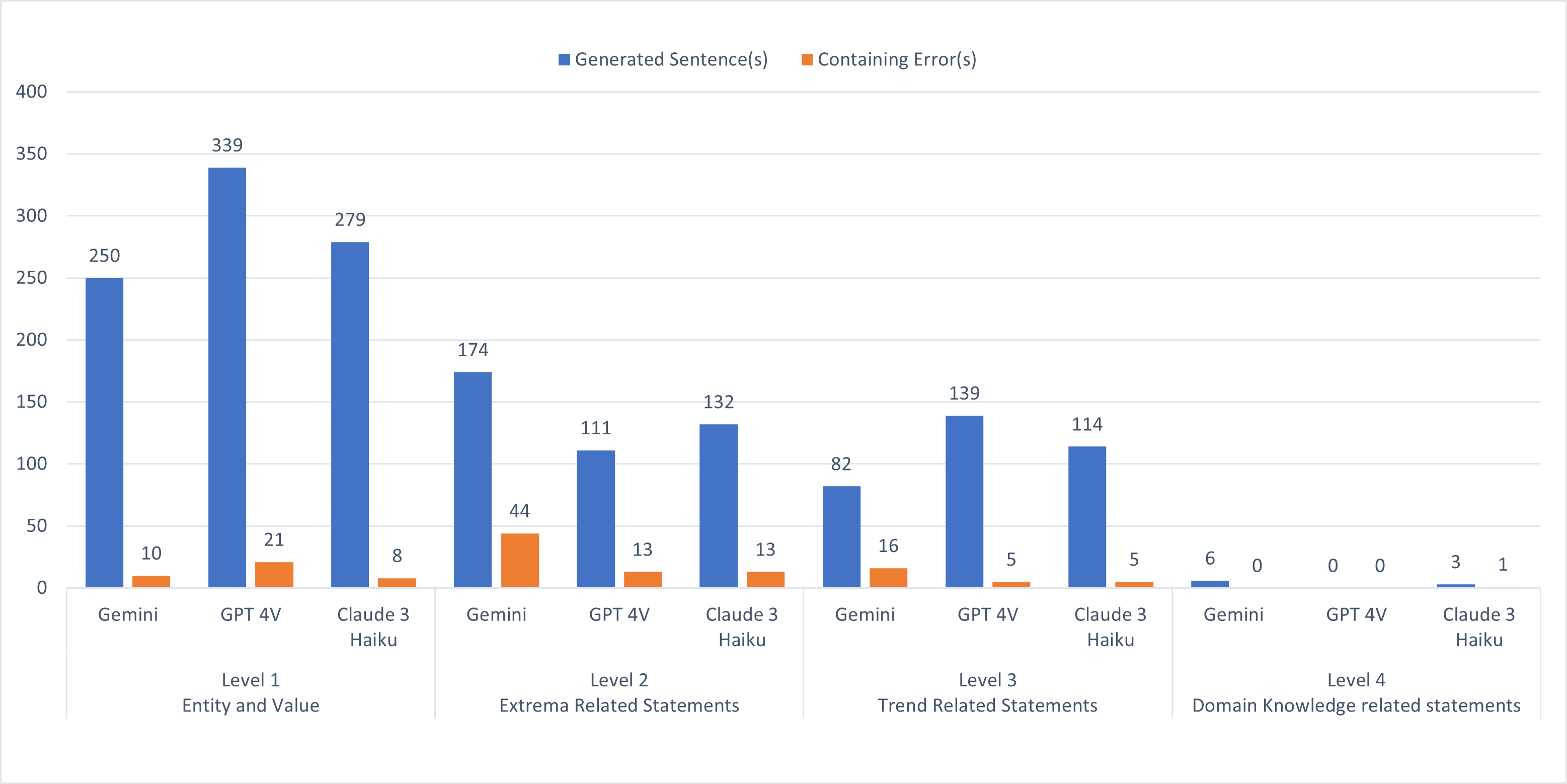}
    \caption{Chart depicts the number of sentences generated by three LVLMs, Gemini, GPT-4V, and Claude-3, at each semantic level (Entity and Value, Extrema Related Statements, Trend Related Statements, Domain Knowledge Related Statements).}
    \label{fig:c2t-four-level-of-semantics}
\end{figure*}

\begin{figure*}[t!]
    \centering
    \includegraphics[width=\textwidth]{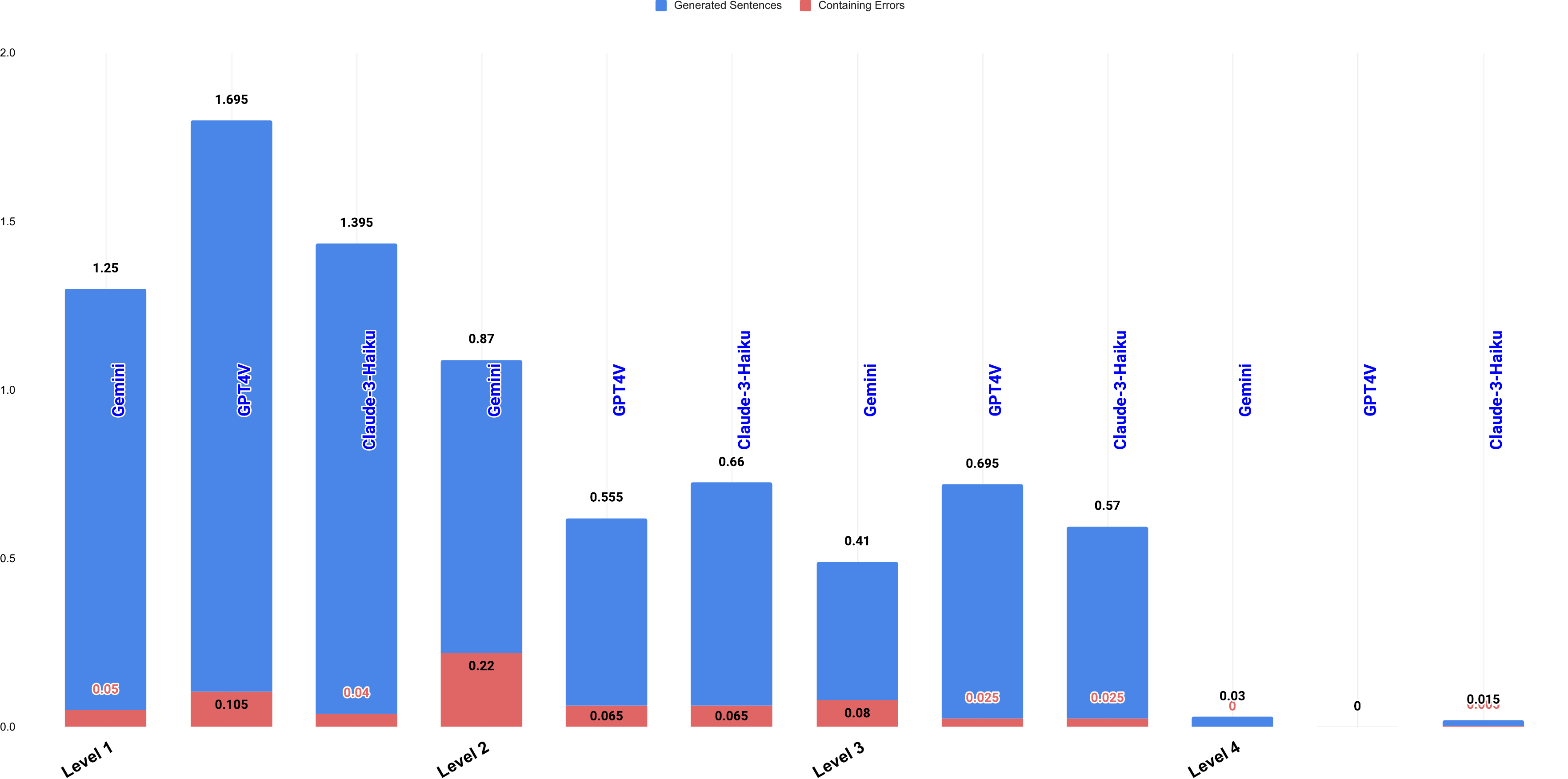}
    \caption{Chart depicts the average number of sentences generated by three LVLMs, Gemini, GPT-4V, and Claude-3, at each semantic level (Entity and Value, Extrema Related Statements, Trend Related Statements, Domain Knowledge Related Statements).}
    \label{fig:c2t-coverage-sem}
\end{figure*}

To assess the quality of summaries generated by Gemini, GPT-4V, and Claude-3, we conducted a detailed analysis of 200 randomly sampled summaries (50 from Pew, 150 from Statista) from the Chart-To-Text dataset generated by these three LVLMs. We meticulously examined each sentence, categorizing it according to the four-level semantic framework. Level-1 sentences focused on entities (axis labels, titles) and chart values. Level-2 highlighted extrema (minimum/maximum) values within the chart, while Level-3 captured trends and patterns. Level 4 addressed sentences requiring domain-specific knowledge external to the chart.

Figure \ref{fig:c2t-four-level-of-semantics} summarizes the key findings. Our analysis revealed the following: \\
In Level-1 (Entity and Value), Gemini generated 250 sentences where 10 sentences contained factual errors; GPT-4V generated 339 sentences, with 21 sentences containing errors; and Claude-3 generated 279 sentences, with 8 sentences containing errors. Both GPT-4V and Claude-3 generated significantly more Level-1 sentences compared to Gemini, with GPT-4V leading in the number of generated sentences but also having a higher error count and percentage (6.19\%).\\
In Level-2 (Extrema Related Statements), Gemini generated 174 sentences, with 44 containing errors; GPT-4V generated 111 sentences, with 13 containing errors; and Claude-3 generated 132 sentences, with 13 containing errors. In this level Gemini produced the most sentences but had a notably higher error rate (25.29\%) compared to GPT-4V (11.71\%) and Claude-3 (9.85\%).\\
In Level-3 (Trend Related Statements), Gemini generated 82 sentences, with 16 containing errors; GPT-4V generated 139 sentences, with 5 containing errors; and Claude-3 generated 114 sentences, with 5 containing errors. GPT-4V significantly outperformed Gemini in Level-3, generating 139 trend-related sentences with only 5 errors, while Claude-3 also performed well with a similar error count but fewer generated sentences. In this level, Gemini still had the highest error percentage of 19.51\% compared to GPT-4V (3.60\%) and Claude-3 (4.39\%).\\
In Level-4 (Domain Knowledge Related Statements), Gemini generated 6 sentences, with no errors; GPT-4V did not generate any sentences; and Claude-3 generated 3 sentences, with 1 containing an error. All three models struggled with Level-4 sentences, reflecting the challenges of incorporating domain-specific knowledge.

\subsubsection{Understanding of 4-level semantics}
\label{app:4-level-sem}
\noindent\textbf{Experimental Setup.}\, In order to evaluate the models in the four semantic levels, we utilize the charts from the ChartQA \cite{masry-etal-2022-chartqa} dataset, and generate 200 different question prompts each for the four semantic levels, i.e., \textit{Level \textbf{1}}, \textit{Level \textbf{2}}, \textit{Level \textbf{3}}, and \textit{Level \textbf{4}}, to evaluate both GPT-4V and Gemini models. Our \textit{Level \textbf{1}} semantic evaluation leveraged a collection of 40 charts encompassing a variety of types. We design five \textit{Level \textbf{1}} questions to assess core aspects of chart construction. These questions targeted attributes such as channel encoding (how data is represented visually), chart type (bar, line, pie, etc.), and axis labeling (x and y). Notably, the chart set comprised a dominant presence of bar charts (70\%), further categorized as horizontal/vertical, simple/stacked/grouped variants. Line charts constituted 17.5\% of the collection, with pie charts making up the remaining 12.5\%. In the case of \textit{Level \textbf{2}}, we design four questions to assess the ability of the models to identify extrema (maxima, minima) and outliers within charts. We include a diverse set of 50 chart types, with bar charts comprising the majority (68\%), followed by line charts (20\%) and pie charts (12\%). In the case of \textit{Level \textbf{3}}, we include a wider range of 100 chart samples, with 28 being line charts. The distribution of chart types at this level remains similar, with bar charts (62\%) holding dominance, followed by line charts (28\%) and pie charts (10\%). Finally, for \textit{Level \textbf{4}}, to evaluate the domain-specific text generation capability of the LVLMs, we employ a test set of 200 distinct chart types. \\
\noindent\textbf{Additional details about the performance of the models.}\, In the `Understanding of 4-level semantics' evaluation, in a subset of \num{40} samples where color encoding information was queried, both GPT-4V and Gemini models struggled. Results indicate that Gemini provided incorrect answers \num{52.5}\% of the time, while GPT-4V had a slightly higher error rate at \num{62.5}\%. In another experiment, for each chart, we designed two questions focused on \textit{Level \textbf{3}} semantic content. We specifically asked \num{56} questions regarding the trends present in the line charts. Our analysis revealed that GPT-4V failed to describe line chart trends correctly in \num{41.07}\% of cases. Gemini demonstrated a higher error rate, failing to identify the correct trend in \num{51.78}\% of instances. While the models excel in recognizing simple, steadily increasing, or decreasing trends in charts related to semantic \textit{Level \textbf{3}}, they struggle with line charts featuring multiple, highly fluctuating lines.

\subsection{Prompt Construction}
\label{app:prompt-const}
In order to come up with the best-performing prompt, we tried many different techniques and used the one that gave a consistent performance. For the zero-shot PAL experiment, we specifically designed the prompt asking the model to output a Python script, which upon execution would give us the final answer to the question. In the case of the 4-Level semantics experiment, we devised questions pertinent to each semantic level and aimed to evaluate the models' proficiency in identifying the various levels of semantic information embedded in the chart image.  We created questions relevant to each of the semantic levels, targeting each of the semantic levels, i.e., \textit{Level - \textbf{1}} (e.g., chart type, x-axis/y-axis labels, color encoding information, etc.), \textit{Level - \textbf{2}} (e.g., maxima, minima, or outliers), \textit{Level - \textbf{3}} (e.g., trends or patterns), \textit{Level - \textbf{4}} (e.g., domain-specific insights). Example prompts can be found in Table \ref{tab:prompt-table}.

\subsection{Additional Experimental Results}
In this section, we present additional experimental results of our automatic evaluation of the Chart-to-text, OpenCQA benchmark, and the newly created ChartQA$^*$ benchmark. Here, Table \ref{tab:chart-to-text-automatic} represents the performance on Chart-to-text, while Table \ref{opencqa-automatic} represents the performance of Gemini and GPT-4V across different metrics, i.e., BLEURT, CIDEr, Perplexity, and BERTScore, and Table \ref{chartqavis-automatic} represents the performance of the models in ChartQA$^*$ benchmark.

\begin{table}[h]
\centering
\caption{Detailed breakdown of LVLMs' performance across chart types in the VisText dataset. Here, `L1', `L2L3' denote the chart caption types.}
\label{tab:vst-cytpe-perf-analysis}
\resizebox{\columnwidth}{!}{%
\begin{tabular}{lcccccc}
\textbf{Model} & \textbf{Area (L1)} & \textbf{Area (L2L3)} & \textbf{Bar (L1)} & \textbf{Bar (L2L3)} & \textbf{Line (L1)} & \textbf{Line (L2L3)} \\\Xhline{1pt}
Gemini            & 33.80 & 18.30 & 30.50 & 17.60 & 33.30 & 19.10 \\
GPT4V             & 21.60 & 13.20 & 21.30 & 12.90 & 21.50 & 14.10 \\
Claude-3-Haiku    & 31.10 & 17.20 & 29.40 & 16.20 & 30.60 & 17.30 \\
Phi-3-vision-inst & 22.80 & 11.90 & 23.80 & 12.10 & 23.60 & 13.10 \\\Xhline{1pt}
\end{tabular}%
}
\end{table}

\begin{table*}[h]
\scriptsize
\centering
\caption{Example of the prompts used to evaluate the LVLMs.}
\label{tab:prompt-table}
\begin{tabularx}{\textwidth}{>{\hsize=0.3\hsize}X|>{\hsize=0.3\hsize}X>{\hsize=1.0\hsize}X}
\Xhline{4\arrayrulewidth}
\textbf{Task} & \textbf{Setup} & \textbf{Prompt} \\ \hline
\midrule
\multirow{2}{=}{ChartQA} & Chain-of-Thought (CoT)  & Given the chart image and a question in the input, generate an appropriate response to the question. Input: \texttt{\{question\}}. Output: Let's think step by step. \\
\cmidrule{2-3}
& Program-aided Language Modeling (PAL) & You will be provided with a chart image and a question associated with it in the input. Create a Python script that, upon execution, generates an answer to the input question. The script should directly incorporate all necessary data, avoiding any supplemental comments or superfluous variables. Ensure that the data is structured within the script to facilitate the calculation. Exclude any extraneous text or explanation after the python script. Output should only contain the python code. Input: \texttt{\{question\}} \\
\midrule
\multirow{4}{=}{4-level of semantic contents} & Level - 1 & \makecell[l]{1. What is the chart type in the input image? \\2. What is the range of x-axis? \\ 3. What is the range of y-axis? \\4. What are the x-axis and y-axis labels in the chart? \\5. What do each of the colors represent in the chart? \\6. What is the chart type in the input image?}  \\
\cmidrule{2-3}
& Level - 2 & \makecell[l]{1. Identify the axis that contains a numerical range. What is \\the maximum value in that axis? \\ 2. Identify the axis that contains a numerical range. What is \\the minimum value in that axis? \\ 3. Are there any outliers in the chart? \\ 4. Compare between the labels that hold the minimum and \\ maximum values.}    \\
\cmidrule{2-3}
& Level - 3 & \makecell[l]{1. What type of trend can you infer from the chart? \\2. Describe the trend that is visible in the chart and provide \\evidence for your conclusion.}  \\
\cmidrule{2-3}
& Level - 4 & Analyze the chart given in the input in one paragraph. \\
\midrule
Open-ended Chart QA & - & Provide an open-ended answer to the following question based on the provided chart image.\\
\midrule
\multirow{3}{=}{Chart Summarization} & Chart-To-Text & The attached chart shows \texttt{\{title\}}. Summarize the chart in a single paragraph 
focusing on trends and important data points. While summarizing, focus on the axis and color-related information in 
the chart. \\
\cmidrule{2-3}
& Vistext (L1) & Summarize the attached chart in a single paragraph focusing on the chart's elemental and encoded properties.\\
\cmidrule{2-3}
& Vistext (L2/L3) & Summarize the attached chart in a single paragraph focusing trends and statistics about the chart. \\
\midrule
Fact-Checking with Charts & - & For the given chart image, determine if the following claim statement in the input is supported by the chart. If supported, then output 'supports', otherwise output 'refutes'. Input: \texttt{\{claim\}} \\
\midrule
Chart-to-Table & - & Extract the underlying data table from the provided chart image. Each row should be on a separate line and use | to separate the cells in the same row by following this format:
Column Header 1 | Column Header 2 \newline
Cell 1 | Cell 2 \newline
Cell 3 | Cell 4 \\
\hline
\end{tabularx}
\end{table*}

\begin{table*}[t]
\centering
\caption{An example from the ChartQA-human test set in the zero-shot CoT setup where both the Gemini-pro-vision and GPT-4V models responded correctly, marked in \textcolor{green}{\textbf{Green}} text.}
\label{tab:pos-ex-CoT}
\scalebox{0.85}{
\begin{tabular}{p{0.7\linewidth}p{0.4\linewidth}}
\Xhline{4\arrayrulewidth}
\multicolumn{1}{l}{\textbf{ChartQA-Human}} & \\ \hline \hline
\multicolumn{1}{p{0.7\linewidth}}{\textbf{Prompt:} Given the chart image and a question in the input, generate an appropriate response to the question. Input: Does the sum of smallest two segment value is greater then the the largest segment value of the graph?  Output: Let's think step by step.} & 
\raisebox{-\height}{\includegraphics[width=0.4\textwidth]{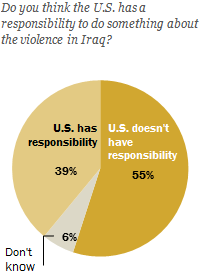}} \\ \hline
\multicolumn{2}{l}{\textbf{Gold-Label:} No} \\
\multicolumn{2}{p{\linewidth}}{\textbf{Gemini-pro-vision Response:}   The smallest two segments are ``Don't know'' and ``U.S. has responsibility''. Their values are 6\% and 39\%, respectively. Their sum is 45\%. The largest segment is ``U.S. doesn't have responsibility'' with the value of 55\%. \textcolor{green}{\textbf{Therefore, the sum of the smallest two segments is not greater than the largest segment value.}}} \\
\multicolumn{2}{p{\linewidth}}{\textbf{GPT-4V Response:} To determine this, we must add the values of the two smallest segments and compare the sum to the value of the largest segment. The smallest segments are 6\% (``Don't know'') and 39\% (``U.S. has responsibility''), which sum up to 45\%. The largest segment is 55\% (``U.S. doesn't have responsibility''). Comparing the two sums, 45\% is less than 55\%, \textcolor{green}{\textbf{meaning the sum of the smallest two segments is not greater than the value of the largest segment.}}} \\ \hline
\end{tabular}
}
\end{table*}
\begin{table*}[t]
\centering
\caption{An example from the ChartQA-human test set in the zero-shot CoT setup where both the Gemini-pro-vision and GPT-4V models responded incorrectly, marked in \textcolor{red}{\textbf{Red}} text.}
\label{tab:neg-ex-CoT}
\scalebox{0.9}{
\begin{tabular}{p{0.6\linewidth}p{0.4\linewidth}}
\Xhline{4\arrayrulewidth}
\multicolumn{1}{l}{\textbf{ChartQA-Human}} & \\ \hline \hline
\multicolumn{1}{p{0.6\linewidth}}{\textbf{Prompt:} Given the chart image and a question in the input, generate an appropriate response to the question. Input: What is the average of the smallest gray bar and largest light blue bar?  Output: Let's think step by step.} & 
\raisebox{-\height}{\includegraphics[width=0.4\textwidth]{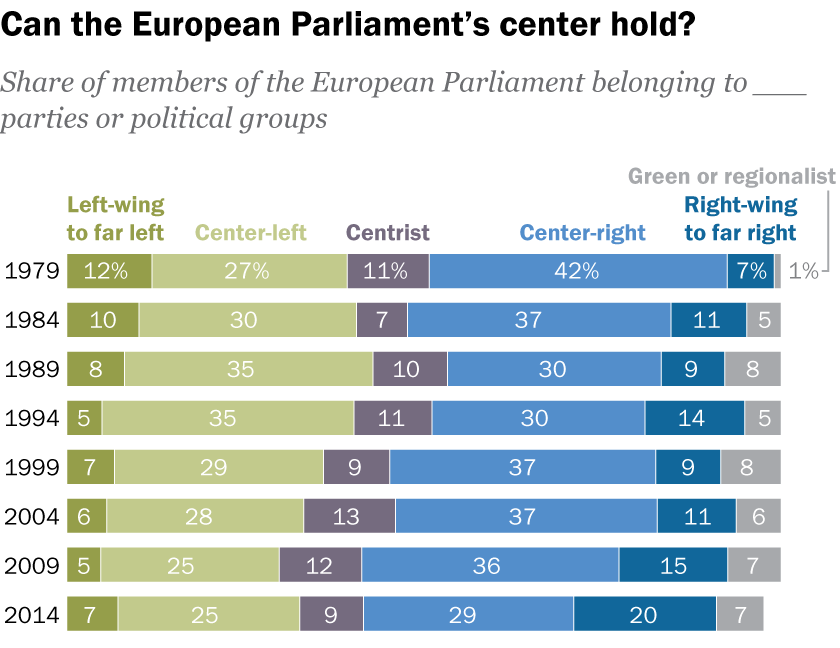}} \\ \hline
\multicolumn{2}{l}{\textbf{Gold-Label:} 21.5} \\
\multicolumn{2}{p{\linewidth}}{\textbf{Gemini-pro-vision Response:} The smallest gray bar is 1\% in 1979 and the largest light blue bar is 37\% in 2004. \textcolor{red}{\textbf{So the average of the two is (1+37) / 2 = 19\%.}}} \\
\multicolumn{2}{p{\linewidth}}{\textbf{GPT-4V Response:} To find the average of the smallest gray bar and the largest light blue bar, we must first identify these values from the chart. In this chart, the gray bars represent the ``Centrist'' political group while the light blue bars represent the ``Center-right'' political group. The smallest percentage for the gray ``Centrist'' bars is in the year 2014 with 9\%. The largest percentage for the light blue ``Center-right'' bars is in the year 2014 with 29\%. Now, we calculate the average of these two percentages:
Average = (Smallest gray bar value + Largest light blue bar value) / 2
Average = (9\% + 29\%) / 2
Average = 38\% / 2
Average = 19\%
\textcolor{red}{\textbf{So, the average of the smallest gray bar and the largest light blue bar is 19\%.}}} \\ \hline
\end{tabular}
}
\end{table*}
\begin{table*}[t]
\centering
\caption{An example from the ChartQA-augmented test set in the zero-shot CoT setup where both the Gemini-pro-vision and GPT-4V models responded correctly, marked in \textcolor{green}{\textbf{Green}} text.}
\label{tab:pos-ex-aug-CoT}
\scalebox{0.92}{
\begin{tabular}{p{0.6\linewidth}p{0.4\linewidth}}
\Xhline{4\arrayrulewidth}
\multicolumn{1}{l}{\textbf{ChartQA-Augmented}} & \\ \hline \hline
\multicolumn{1}{p{0.6\linewidth}}{\textbf{Prompt:} Given the chart image and a question in the input, generate an appropriate response to the question. Input: What was the average number of applications per job posting on LinkedIn in the United States?  Output: Let's think step by step.} & 
\raisebox{-\height}{\includegraphics[width=0.4\textwidth]{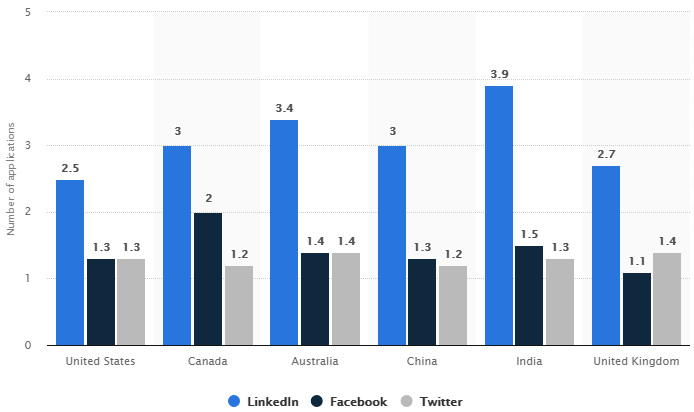}} \\ \hline
\multicolumn{2}{l}{\textbf{Gold-Label:} 2.5} \\
\multicolumn{2}{p{\linewidth}}{\textbf{Gemini-pro-vision Response:}   \textcolor{green}{\textbf{The average number of applications per job posting on LinkedIn in the United States was 2.5.}}} \\
\multicolumn{2}{p{\linewidth}}{\textbf{GPT-4V Response:} \textcolor{green}{\textbf{Based on the chart provided, the average number of applications per job posting on LinkedIn in the United States is 2.5.}}} \\ \hline
\end{tabular}
}
\end{table*}
\begin{table*}[t]
\centering
\caption{An example from the ChartQA-augmented test set in the zero-shot CoT setup where both the Gemini-pro-vision and GPT-4V models responded incorrectly, marked in \textcolor{red}{\textbf{Red}} text.}
\label{tab:neg-ex-aug-CoT}
\scalebox{0.85}{
\begin{tabular}{p{0.6\linewidth}p{0.4\linewidth}}
\Xhline{4\arrayrulewidth}
\multicolumn{1}{l}{\textbf{ChartQA-Augmented}} & \\ \hline \hline
\multicolumn{1}{p{0.7\linewidth}}{\textbf{Prompt:} Given the chart image and a question in the input, generate an appropriate response to the question. Input: In what year did the coverage ratio for bad loans increase steadily?  Output: Let's think step by step.} & 
\raisebox{-\height}{\includegraphics[width=0.4\textwidth]{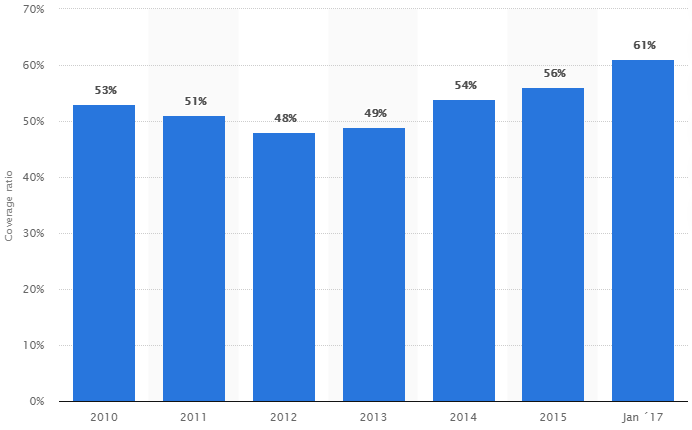}} \\ \hline
\multicolumn{2}{l}{\textbf{Gold-Label:} 21.5} \\
\multicolumn{2}{p{\linewidth}}{\textbf{Gemini-pro-vision Response:} \textcolor{red}{\textbf{The coverage ratio for bad loans increased steadily from 2010 to 2017.}} In 2010, the coverage ratio was 53\%. In 2011, it increased to 51\%. In 2012, it increased to 48\%. In 2013, it increased to 49\%. In 2014, it increased to 54\%. In 2015, it increased to 56\%. In 2016, it increased to 61\%.} \\
\multicolumn{2}{p{\linewidth}}{\textbf{GPT-4V Response:} \textcolor{red}{\textbf{The coverage ratio for bad loans began to increase steadily starting in the year 2013.}} From this point onward, the chart shows a consistent upward trend, from 48\% in the year 2012 to 61\% in January 2017.} \\ \hline
\end{tabular}
}
\end{table*}
\begin{table*}[t]
\centering
\caption{An example from the ChartQA-human test set in the zero-shot PAL setup where both the Gemini-pro-vision and GPT-4V models responded correctly, marked in \textcolor{green}{\textbf{Green}} text.}
\label{tab:pos-ex-hum-PAL}
\scalebox{0.85}{
\begin{tabular}{p{0.7\linewidth}p{0.4\linewidth}}
\Xhline{4\arrayrulewidth}
\multicolumn{1}{l}{\textbf{ChartQA-Human}} & \\ \hline \hline
\multicolumn{1}{p{0.7\linewidth}}{\textbf{Prompt:} You will be provided with a chart image and a question associated with it in the input. Create a Python script that, upon execution, generates an answer to the input question. The script should directly incorporate all necessary data, avoiding any supplemental comments or superfluous variables. Ensure that the data is structured within the script to facilitate the calculation. Exclude any extraneous text or explanation after the python script. Output should only contain the python code. Input: What's the ratio of smallest segment and second largest segment?} & 
\raisebox{-\height}{\includegraphics[width=0.4\textwidth]{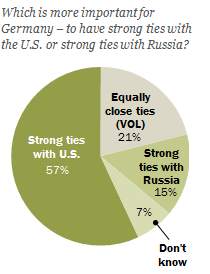}} \\ \hline
\multicolumn{2}{l}{\textbf{Gold-Label:} 0.33333333} \\
\multicolumn{2}{p{\linewidth}}{\textbf{Gemini-pro-vision Response:}   \textcolor{green}{\textbf{Ratio of smallest (7) and second largest (21) segments: 0.33.}}} \\
\multicolumn{2}{p{\linewidth}}{\textbf{GPT-4V Response:} \textcolor{green}{\textbf{0.3333333333}}} \\ \hline
\end{tabular}
}
\end{table*}
\begin{table*}[t]
\centering
\caption{An example from the ChartQA-human test set in the zero-shot PAL setup where both the Gemini-pro-vision and GPT-4V models responded incorrectly, marked in \textcolor{red}{\textbf{Red}} text.}
\label{tab:neg-ex-hum-PAL}
\scalebox{0.9}{
\begin{tabular}{p{0.6\linewidth}p{0.4\linewidth}}
\Xhline{4\arrayrulewidth}
\multicolumn{1}{l}{\textbf{ChartQA-Human}} & \\ \hline \hline
\multicolumn{1}{p{0.6\linewidth}}{\textbf{Prompt:} You will be provided with a chart image and a question associated with it in the input. Create a Python script that, upon execution, generates an answer to the input question. The script should directly incorporate all necessary data, avoiding any supplemental comments or superfluous variables. Ensure that the data is structured within the script to facilitate the calculation. Exclude any extraneous text or explanation after the python script. Output should only contain the python code. Input: What's the median value of red graph?} & 
\raisebox{-\height}{\includegraphics[width=0.4\textwidth]{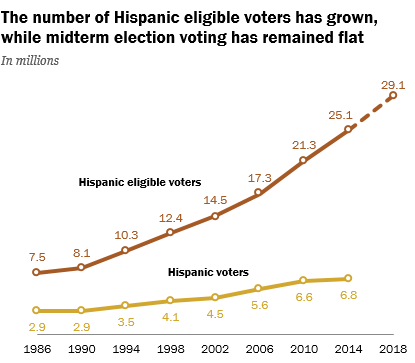}} \\ \hline
\multicolumn{2}{l}{\textbf{Gold-Label:} 14.5} \\
\multicolumn{2}{p{\linewidth}}{\textbf{Gemini-pro-vision Response:} \textcolor{red}{\textbf{4.5}}} \\
\multicolumn{2}{p{\linewidth}}{\textbf{GPT-4V Response:} 
\textcolor{red}{\textbf{4.3}}} \\ \hline
\end{tabular}
}
\end{table*}
\begin{table*}[t]
\centering
\caption{An example from the ChartQA-Augmented test set in the zero-shot PAL setup where both the Gemini-pro-vision and GPT-4V models responded correctly, marked in \textcolor{green}{\textbf{Green}} text.}
\label{tab:pos-ex-aug-PAL}
\scalebox{0.85}{
\begin{tabular}{p{0.7\linewidth}p{0.4\linewidth}}
\Xhline{4\arrayrulewidth}
\multicolumn{1}{l}{\textbf{ChartQA-Augmented}} & \\ \hline \hline
\multicolumn{1}{p{0.7\linewidth}}{\textbf{Prompt:} You will be provided with a chart image and a question associated with it in the input. Create a Python script that, upon execution, generates an answer to the input question. The script should directly incorporate all necessary data, avoiding any supplemental comments or superfluous variables. Ensure that the data is structured within the script to facilitate the calculation. Exclude any extraneous text or explanation after the python script. Output should only contain the python code. Input: What was the share of fish and seafood expenditure of U.S. households in 2018?} & 
\raisebox{-\height}{\includegraphics[width=0.4\textwidth]{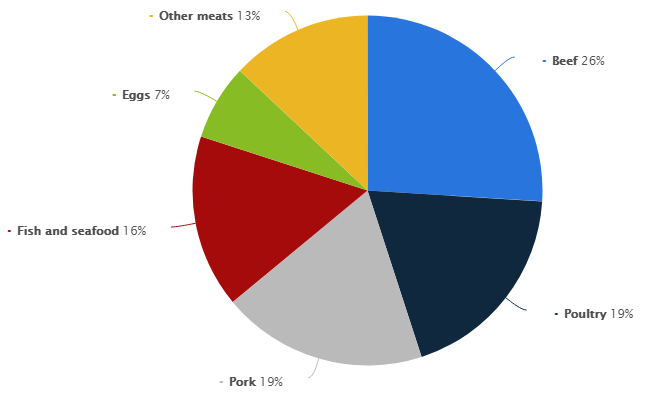}} \\ \hline
\multicolumn{2}{l}{\textbf{Gold-Label:} 16} \\
\multicolumn{2}{p{\linewidth}}{\textbf{Gemini-pro-vision Response:} \textcolor{green}{\textbf{The share of fish and seafood expenditure of U.S. households in 2018 was 16\%}}} \\
\multicolumn{2}{p{\linewidth}}{\textbf{GPT-4V Response:} \textcolor{green}{\textbf{16}}} \\ \hline
\end{tabular}
}
\end{table*}
\begin{table*}[t]
\centering
\caption{An example from the ChartQA-augmented test set in the zero-shot PAL setup where both the Gemini-pro-vision and GPT-4V models responded incorrectly, marked in \textcolor{red}{\textbf{Red}} text.}
\label{tab:neg-ex-aug-PAL}
\scalebox{0.85}{
\begin{tabular}{p{0.6\linewidth}p{0.4\linewidth}}
\Xhline{4\arrayrulewidth}
\multicolumn{1}{l}{\textbf{ChartQA-Augmented}} & \\ \hline \hline
\multicolumn{1}{p{0.7\linewidth}}{\textbf{Prompt:} You will be provided with a chart image and a question associated with it in the input. Create a Python script that, upon execution, generates an answer to the input question. The script should directly incorporate all necessary data, avoiding any supplemental comments or superfluous variables. Ensure that the data is structured within the script to facilitate the calculation. Exclude any extraneous text or explanation after the python script. Output should only contain the python code. Input: How many cases of COVID-19 were confirmed across eight hospitals in Tokyo?} & 
\raisebox{-\height}{\includegraphics[width=0.4\textwidth]{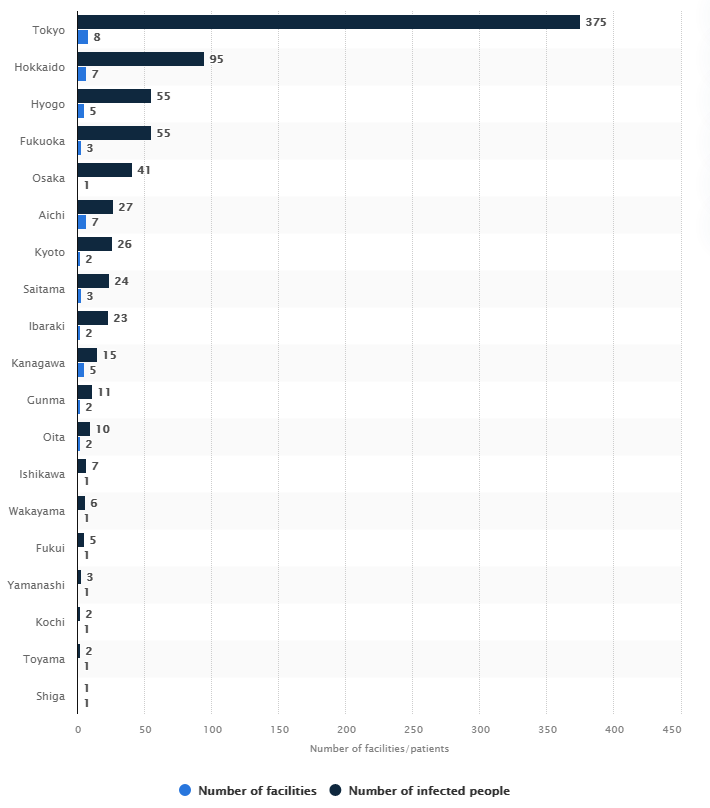}} \\ \hline
\multicolumn{2}{l}{\textbf{Gold-Label:} 375} \\
\multicolumn{2}{p{\linewidth}}{\textbf{Gemini-pro-vision Response:} \textcolor{red}{\textbf{The total number of cases across the eight hospitals in Tokyo is 666.}}} \\
\multicolumn{2}{p{\linewidth}}{\textbf{GPT-4V Response:} \textcolor{red}{\textbf{3000}}} \\ \hline
\end{tabular}
}
\end{table*}
\begin{table*}[t]
\centering
\caption{An example from the Chart-to-Text Pew test set. Factually incorrect summaries, marked in \textcolor{red}{Red} text.}
\label{tab:app-c2t-pew}
\scalebox{0.85}{
\begin{tabular}{p{0.6\linewidth}p{0.4\linewidth}}
\Xhline{4\arrayrulewidth}
\multicolumn{1}{l}{\textbf{Chart-to-Text Pew}} & \\ \hline \hline
\multicolumn{1}{p{0.7\linewidth}}{\textbf{Prompt:} The attached chart shows Following Election News Very Closely Democrat. Summarize the chart in a single paragraph focusing on trends and important data points. While summarizing, focus on the axis and color-related information in the chart.} & 
\raisebox{-\height}{\includegraphics[width=0.4\textwidth]{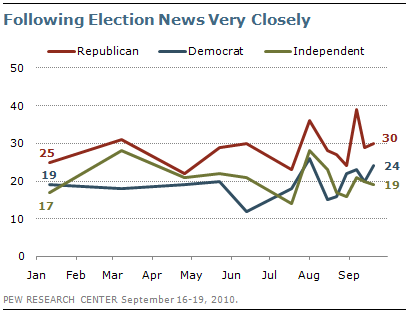}} \\ \hline
\multicolumn{2}{p{\linewidth}}{\textbf{Gold-Summary:} While election news dominated media coverage last week, the public continued to show relatively modest interest. According to a separate analysis by the Pew Research Center’s Project for Excellence in Journalism, midterm news made up 30\% of the newshole. Just more than one-in-ten (11\%) say this was the news they followed most closely. Nearly a quarter (23\%) say they followed election news very closely.
Three-in-ten Republicans say they followed election news very closely last week, not much different from the 24\% of Democrats that say this. Two-in-ten independents (19\%) say they followed election news very closely.} \\
\multicolumn{2}{p{\linewidth}}{\textbf{Gemini-pro-vision Response:}   The chart shows the percentage of Republicans, Democrats, and Independents who are following election news very closely. The x-axis shows the time, and the y-axis shows the percentage. The data is from January to September 2010. The highest percentage of Republicans who were following election news very closely was in \textcolor{red}{September (44\%)}. The lowest percentage of Republicans who were following election news very closely was in \textcolor{red}{February (25\%)}. The highest percentage of Democrats who were following election news very closely was in \textcolor{red}{September (30\%)}. The lowest percentage of Democrats who were following election news very closely was in \textcolor{red}{May (17\%)}. The highest percentage of Independents who were following election news very closely was in \textcolor{red}{September (24\%)}. The lowest percentage of Independents who were following election news very closely was in \textcolor{red}{February (14\%)}.} \\
\multicolumn{2}{p{\linewidth}}{\textbf{GPT-4V Response:} The chart visualizes the percentage of Republicans (in red), Democrats (in blue), and Independents (in green) who followed election news very closely over a period from January to September 2011. Starting in January, \textcolor{red}{Democrats showed a lower percentage (17\%)} compared to Republicans (25\%) and \textcolor{red}{Independents (19\%)}. Through the months, all three groups experienced fluctuations in the level of attention paid to election news. Notable peaks can be observed for Republicans around \textcolor{red}{May} and September (both at 30\%) and for Democrats in May (around 24\%) and July, with a subsequent drop and a rise back to around 24\% in September. Independent interest peaked in August, close to 30\%. Overall, a trend can be seen with Republicans consistently showing the highest percentage of individuals following the election news very closely, with Democrats and Independents generally trailing, albeit with significant increases toward the latter part of the surveyed period.} \\ 
\multicolumn{2}{p\linewidth}{\textbf{Claude-3-Haiku: } The chart shows the percentage of people from different political affiliations who reported following election news very closely over time. \textcolor{red}{The Democratic line starts high in January at around The Ind 25\% and then fluctuates between 15-30\% throughout the year. The Republican line starts lower in January around 19\% but then rises to a peak of around 40\% in September.ependent line remains relatively flat, staying between 15-25\% for most of the year.} The chart highlights the increased attention to election news among Republicans towards the end of the year compared to Democrats and Independents.} \\ \hline
\end{tabular}
}
\end{table*}
\begin{table*}[t]
\centering
\caption{An example from the Chart-to-Text Statista test set. Factually incorrect summaries, marked in \textcolor{red}{Red} text.}
\label{tab:app-c2t-statista}
\scalebox{0.85}{
\begin{tabular}{p{0.6\linewidth}p{0.4\linewidth}}
\Xhline{4\arrayrulewidth}
\multicolumn{1}{l}{\textbf{Chart-to-Text Statista}} & \\ \hline \hline
\multicolumn{1}{p{0.7\linewidth}}{\textbf{Prompt:} The attached chart shows United Kingdom: Population growth from 2009 to 2019 (compared to the previous year). Summarize the chart in a single paragraph focusing on trends and important data points. While summarizing, focus on the axis and color-related information in the chart.} & 
\raisebox{-\height}{\includegraphics[width=0.4\textwidth]{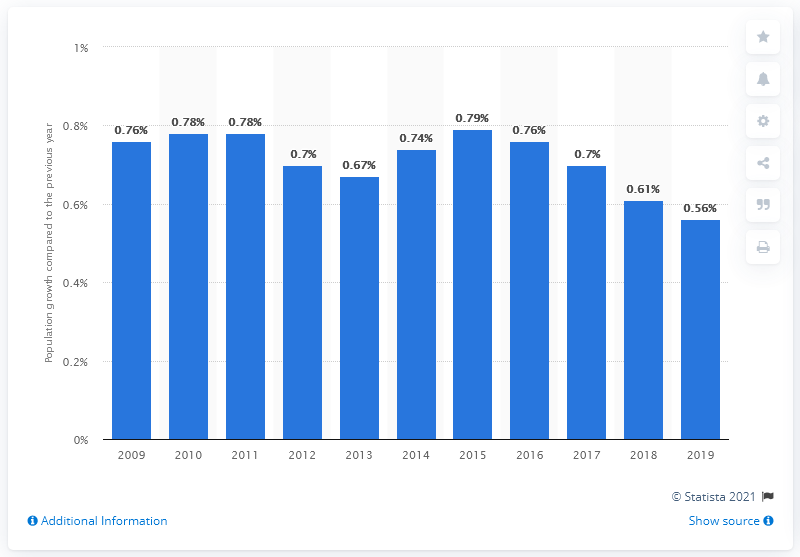}} \\ \hline
\multicolumn{2}{p{\linewidth}}{\textbf{Gold-Summary:} This statistic shows the population growth in the United Kingdom from 2009 to 2019. In 2019, the UK population grew by about 0.56 percent compared to the previous year.} \\

\multicolumn{2}{p{\linewidth}}{\textbf{Gemini-pro-vision Response:}   The line chart shows the United Kingdom's population growth from 2009 to 2019. \textcolor{red}{The population growth rate was highest in 2011 at 0.78\%}, followed by \textcolor{red}{2015 and 2016 at 0.76\%}. The lowest population growth rate was in 2019 at 0.56\%. The population growth rate decreased from 2011 to 2013, increased from 2013 to \textcolor{red}{2016}, and then decreased again from 2016 to 2019.} \\

\multicolumn{2}{p{\linewidth}}{\textbf{GPT-4V Response:} The chart depicts the population growth rate of the United Kingdom from 2009 to 2019 as compared to the previous year. The vertical axis indicates the growth rate percentage, while the horizontal axis lists the years. The bars are colored in shades of blue, with a darker blue representing each year's growth rate. From 2009, there's a slight fluctuation in the growth rate, with it peaking in \textcolor{red}{2014 at 0.79\%}, indicating that year experienced the highest growth relative to the prior year within this timeframe. Overall, the rate starts at 0.76\% in 2009, fluctuates slightly but stays generally consistent through 2014, and then exhibits a declining trend from 2015 onwards, reaching a low of 0.56\% in 2019. The chart shows a gradual reduction in population growth rates in the latter part of the decade, implying a slowdown in the rate at which the population of the UK is increasing.} \\ 
\multicolumn{2}{p\linewidth}{\textbf{Claude-3-Haiku:} The chart shows the population growth rate in the United Kingdom from 2009 to 2019, compared to the previous year. The y-axis represents the population growth rate, expressed as a percentage. The data points reveal a general downward trend in the population growth rate over the 10-year period, declining from 0.76\% in 2009 to 0.56\% in 2019. The highest growth rate was observed in 2015 at 0.79\%, while the \textcolor{red}{lowest was in 2013 at 0.67\%}. The chart provides a clear visual representation of the changes in the UK's population growth over the last decade.} \\ \hline
\end{tabular}
}
\end{table*}
\begin{table*}[t]
\centering
\caption{An example from the Chart-to-Text Statista test set, where summary generated by GPT-4V contains visually incorrect information (mentioning that there are multiple shades of blue color, whereas there are no shades present in the chart), marked in \textcolor{blue}{Blue} text, while Gemini-pro-vision \& Claude-3-Haiku didn't include in visual information.}
\label{tab:app-c2t-visual-incorrect}
\scalebox{0.85}{
\begin{tabular}{p{0.6\linewidth}p{0.4\linewidth}}
\Xhline{4\arrayrulewidth}
\multicolumn{1}{l}{\textbf{Chart-to-Text}} & \\ \hline \hline
\multicolumn{1}{p{0.7\linewidth}}{\textbf{Prompt:} The attached chart shows global market share of smartphones with a triple camera setup in 2019, by brand. Summarize the chart in a single paragraph focusing on trends and important data points. While summarizing, focus on the axis and color-related information in the chart.} & 
\raisebox{-\height}{\includegraphics[width=0.4\textwidth]{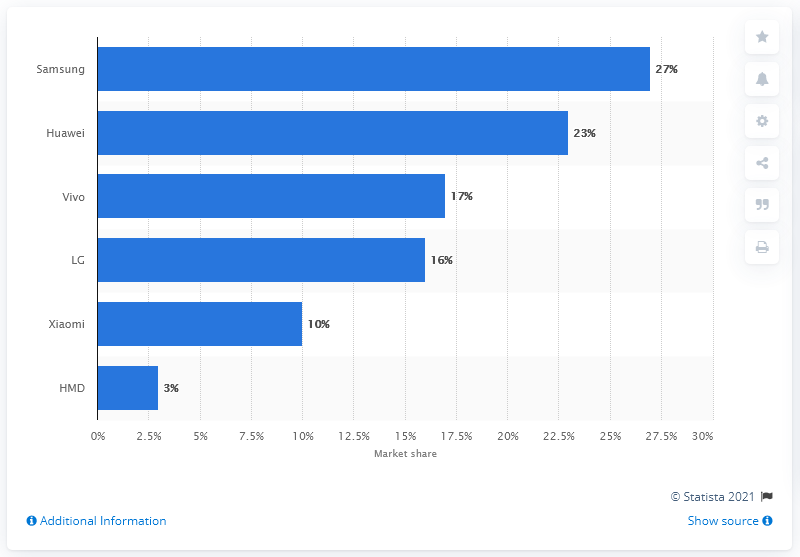}} \\ \hline
\multicolumn{2}{p{\linewidth}}{\textbf{Gold-Summary:} The ranking illustrates the global market share of smartphones with a triple camera setup in 2019, by brand. In 2019, Samsung accounted for 27 percent of the triple camera smartphone market.
} \\

\multicolumn{2}{p{\linewidth}}{\textbf{Gemini-pro-vision Response:}   The chart shows the global market share of smartphones with a triple camera setup in 2019, by brand. Samsung is the market leader with a 27\% share. Huawei is second with a 23\% share. Vivo is third with a 17\% share. LG is fourth with a 16\% share. Xiaomi is fifth with a 10\% share. HMD is sixth with a 3\% share. The chart shows that Samsung and Huawei are the clear leaders in the market for smartphones with a triple camera setup. Vivo, LG, and Xiaomi are also significant players in the market. HMD is a distant sixth.} \\

\multicolumn{2}{p{\linewidth}}{\textbf{GPT-4V Response:} The chart illustrates the global market share of smartphones featuring a triple camera setup in 2019, segmented by brand. \textcolor{blue}{The horizontal bars, shaded in varying shades of blue, represent each brand's market share}, displayed on the x-axis which ranges from 0\% to 30\%. Samsung leads the market with the highest share, represented by the \textcolor{blue}{darkest shade of blue}, accounting for 27\% of the market. Huawei follows closely, with a 23\% share in a \textcolor{blue}{slightly lighter blue}. Vivo and LG come next with 17\% and 16\%, respectively; their \textcolor{blue}{corresponding bars are in progressively lighter shades of blue}. Xiaomi holds a 10\% share, while HMD has the smallest segment visualized, at 3\%, \textcolor{blue}{depicted with the lightest blue}. The chart is a visual representation of the significant hold Samsung and Huawei have in the triple camera smartphone segment, while also showing a quite notable presence of other brands such as Vivo, LG, and Xiaomi in this market.} \\ 
\multicolumn{2}{p\linewidth}{\textbf{Claude-3-Haiku: } The chart shows the global market share of smartphones with a triple camera setup in 2019, broken down by brand. Samsung leads the market with a 27\% share, followed by Huawei at 23\%. Vivo and LG have lower market shares of 17\% and 16\%, respectively. Xiaomi holds a 10\% market share, while HMD (the company behind Nokia phones) has the smallest share at 3\%. The chart presents the data in a clear and visually appealing manner, with the market share percentages displayed directly on the bars for each brand.} \\\hline
\end{tabular}
}
\end{table*}
\begin{table*}[t]
\centering
\caption{An example from the Chart-to-Text Statista test set, where summaries generated by GPT-4V and Gemini-pro-vision captured trend correctly, marked in \textcolor{green}{Green} text.}
\label{tab:app-c2t-correct-trend}
\scalebox{0.85}{
\begin{tabular}{p{0.6\linewidth}p{0.4\linewidth}}
\Xhline{4\arrayrulewidth}
\multicolumn{1}{l}{\textbf{Chart-to-Text}} & \\ \hline \hline
\multicolumn{1}{p{0.7\linewidth}}{\textbf{Prompt:} The attached chart shows HeidelbergCement's global revenue from FY 2009 to FY 2018 (in billion euros). Summarize the chart in a single paragraph focusing on trends and important data points. While summarizing, focus on the axis and color-related information in the chart.} & 
\raisebox{-\height}{\includegraphics[width=0.4\textwidth]{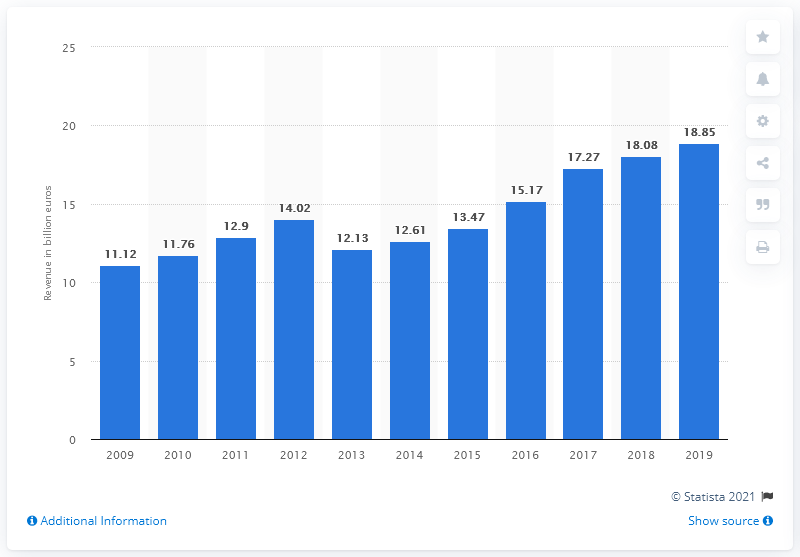}} \\ \hline
\multicolumn{2}{p{\linewidth}}{\textbf{Gold-Summary:} This statistic represents HeidelbergCement's global revenue between the fiscal year of 2009 and the fiscal year of 2019. In the fiscal year of 2019, the German multinational building materials company generated around 18.85 billion euros (or about 21.1 billion U.S. dollars) in revenue.
} \\

\multicolumn{2}{p{\linewidth}}{\textbf{Gemini-pro-vision Response:}  \textcolor{green}{HeidelbergCement's global revenue experienced fluctuations but showed an overall increase from FY 2009 to FY 2018.} The revenue started at 11.12 billion euros in FY 2009 and reached 18.85 billion euros in FY 2018, representing a growth of 70\%. The highest increase occurred from FY 2016 to FY 2017, when the revenue jumped from 15.17 billion euros to 17.27 billion euros. The color blue is used to represent the data in the chart.} \\

\multicolumn{2}{p{\linewidth}}{\textbf{GPT-4V Response:} The chart shows HeidelbergCement's global revenue from the fiscal year 2009 to 2018, with revenue figures represented in billions of euros on the vertical axis and the fiscal years on the horizontal axis. The chart displays \textcolor{green}{a positive trend in revenue over the ten-year period, with some fluctuations}. Starting at €11.12 billion in 2009, there was a slight increase until 2012 when revenue peaked at €14.02 billion. After a decrease to €13.47 billion in 2015, the revenue then showed a significant upward trend, reaching the highest point of €17.27 billion in 2017 before a slight decrease in 2018 to €18.08 billion. The bars in the chart are uniformly colored in blue, and the year-over-year changes are clearly indicated by the varying heights of these bars. The upward trend signifies a general growth in HeidelbergCement's global revenue over the decade.} \\ 
\multicolumn{2}{p\linewidth}{\textbf{Claude-3-Haiku: } The chart shows the global revenue of HeidelbergCement from 2009 to 2019, measured in billions of euros. \textcolor{green}{The revenue has steadily increased over the years}, starting at 11.12 billion euros in 2009 and reaching a high of 18.85 billion euros in 2019. The most significant increase occurred between 2011 and 2012, where the revenue jumped from 12.9 billion euros to 14.02 billion euros. Since then, \textcolor{green}{the revenue has continued to grow, with some fluctuations, but overall maintaining an upward trend.}} \\\hline
\end{tabular}
}
\end{table*}
\begin{table*}[t]
\centering
\caption{An example from the Chart-to-Text Statista test set, where summaries generated by GPT-4V captured trend correctly, marked in \textcolor{green}{Green} text.}
\label{tab:app-c2t-missing-trend}
\scalebox{0.85}{
\begin{tabular}{p{0.6\linewidth}p{0.4\linewidth}}
\Xhline{4\arrayrulewidth}
\multicolumn{1}{l}{\textbf{Chart-to-Text}} & \\ \hline \hline
\multicolumn{1}{p{0.7\linewidth}}{\textbf{Prompt:} The attached chart shows frequency of car models stolen to order in Great Britain (UK) in 2011 and 2012 (in offences per 10,000 registered vehicles). While summarizing, focus on the axis and color-related information in the chart.} & 
\raisebox{-\height}{\includegraphics[width=0.4\textwidth]{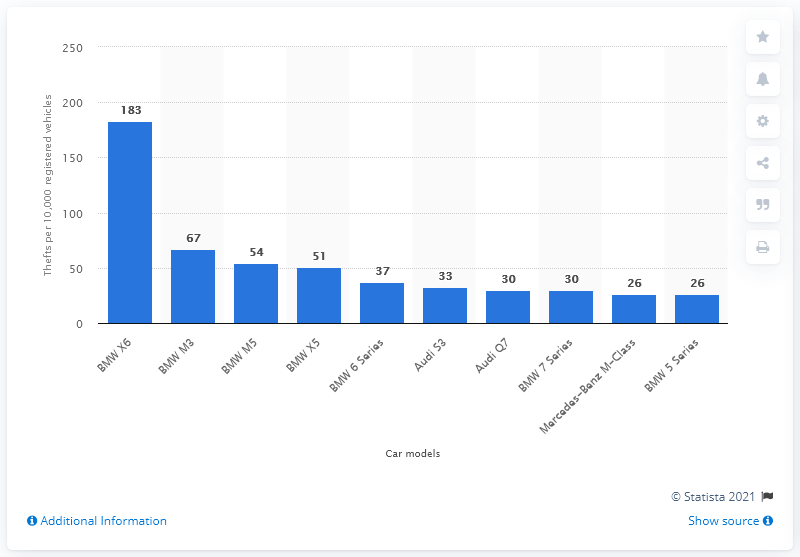}} \\ \hline
\multicolumn{2}{p{\linewidth}}{\textbf{Gold-Summary:} This statistic shows the frequency with which luxury or prestige cars are stolen in Great Britain in 2011 and 2012. These types of luxury cars are mostly stolen to order, meaning that they were sought out specifically on request rather than stolen by an opportunist. BMW, especially the X6 model, was by far the luxury brand which was stolen the most over this period.
} \\

\multicolumn{2}{p{\linewidth}}{\textbf{Gemini-pro-vision Response:} The chart shows the frequency of car models stolen to order in Great Britain (UK) in 2011 and 2012, with the BMW X6, BMW M3, and BMW M5 being the most stolen car models. The BMW X6 had the highest frequency, with 183 thefts per 10,000 registered vehicles. The BMW M3 and BMW M5 had 67 and 54 thefts per 10,000 registered vehicles, respectively. Other car models with significant thefts include the BMW X5, BMW 6 Series, Audi S3, Audi Q7, Mercedes-Benz CLS-Class, and BMW 5 Series, with 51, 37, 33, 30, 30, and 26 thefts per 10,000 registered vehicles, respectively.} \\

\multicolumn{2}{p{\linewidth}}{\textbf{GPT-4V Response:} The attached chart displays the frequency of certain car models being stolen to order in Great Britain for the years 2011 and 2012, with the vertical axis representing the number of thefts per 10,000 registered vehicles and the horizontal axis listing the car models. The BMW X6 stands out as the most targeted model, with a theft rate significantly higher at 183 thefts per 10,000 vehicles. The next most stolen models are the BMW M3 and BMW M5, with lower but still considerable theft rates of 67 and 54 respectively. The frequency then gradually decreases with other BMW models and a few Audi and Mercedes-Benz models, such as the Audi S3 and Mercedes-Benz M-Class showing the lowest frequency of the listed models with 26 thefts each. \textcolor{green}{Overall, the chart indicates a clear preference amongst car thieves for BMW models, specifically the high-performance 'M' series and the X6 SUV.}} \\ 
\multicolumn{2}{p\linewidth}{\textbf{Claude-3-Haiku: } The chart depicts the frequency of car models stolen to order in Great Britain, with the car models displayed on the horizontal axis and the frequency of offenses per 10,000 registered vehicles shown on the vertical axis. The data indicates that the BMW X6 model had the highest frequency of thefts, with 183 offenses per 10,000 registered vehicles. Other models with relatively high theft rates include the BMW 1 Series (67 offenses), BMW 4 Series (54 offenses), and BMW 5 Series (51 offenses). The chart also shows that smaller and more affordable models, such as the Mercedes-Benz A-Class and BMW 3 Series, had relatively low theft rates, with 26 offenses per 10,000 registered vehicles each.} \\\hline
\end{tabular}
}
\end{table*}

\begin{table*}[t]
\centering
\caption{Example errors encountered by the Gemini-vision-pro and the GPT-4V when asked about color encoding information about two different charts. In both cases, the Gemini-pro-vision model did not identify different color encodings at all. In the first figure, the GPT-4V model failed to identify different colors correctly, and in the second figure, the model failed to identify different shades of the same color (in this case \textit{`blue'}) correctly. Erroneous text is marked in \textcolor{red}{\textbf{Red}}.}
\label{tab:color-shade-l1err}
\scalebox{0.95}{
\begin{tabular}{p{0.5\linewidth}p{0.5\linewidth}}
\Xhline{4\arrayrulewidth}
\multicolumn{1}{l}{\textbf{Semantic Evaluation: Level - 1}} & \\ \hline \hline
\multicolumn{1}{p{0.5\linewidth}|}{\raisebox{-\height}{\includegraphics[width=0.5\textwidth]{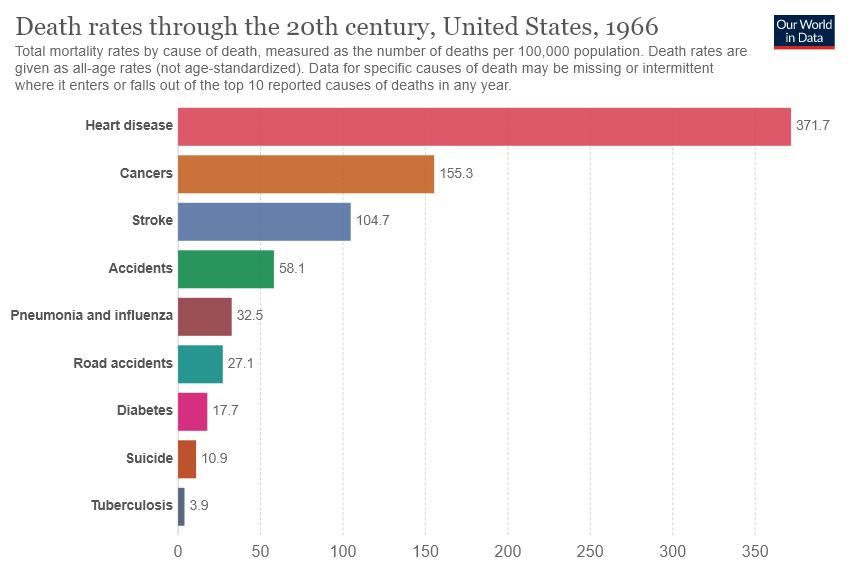}}} &
\textbf{Prompt:} What do each of the colors represent in the chart?
\newline\textbf{Gemini-pro-vision Response:} \textcolor{red}{\textbf{The colors represent the different causes of death.}}
\newline\textbf{GPT-4V Response:} The chart is a horizontal bar graph showing the death rates per 100,000 population from various causes in the United States for the year 1966. Each color on the chart represents a different cause of death, as indicated by the labels on the left side of the graph. Here are the colors and corresponding causes of death they represent:
\newline- Reddish-pink: Heart disease
\newline- Orange: Cancers
\newline- Dark teal blue: Stroke
\newline- Green: Accidents
\newline\textcolor{red}{\textbf{- Light blue: Pneumonia and influenza
\newline- Dark blue: Road accidents}}
\newline- Purple: Diabetes
\newline\textcolor{red}{\textbf{- Yellow: Suicide
\newline- Light green: Tuberculosis}}
\newline Adjacent to each color bar is the specific death rate for that cause of death per 100,000 population, with heart disease being the leading cause at 371.7 deaths per 100,000 and tuberculosis being the least within the top causes of death listed at 3.9 deaths per 100,000.
 \\ \midrule
\multicolumn{1}{p{0.5\linewidth}|}{\raisebox{-\height}{\includegraphics[width=0.5\textwidth]{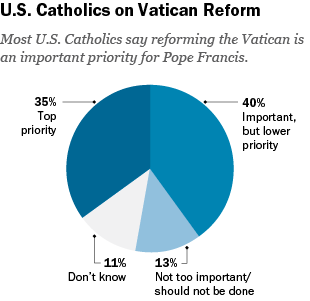}}} &
\textbf{Prompt:} What do each of the colors represent in the chart?
\newline\textbf{Gemini-pro-vision Response:} The blue color represents the percentage of respondents who think that reforming the Vatican is an important priority for Pope Francis. \textcolor{red}{\textbf{The white color represents the percentage of respondents who think that it is not too important or should not be done.}} The gray color represents the percentage of respondents who don't know.
\newline\textbf{GPT-4V Response:} The chart is a pie chart showing the opinions of U.S. Catholics on the importance of reforming the Vatican as a priority for Pope Francis, based on the information provided by the Pew Research Center. Here's what each color represents:
\newline
- \textcolor{red}{\textbf{Light Blue: 35\% view reforming the Vatican as a top priority.}}
\newline
- \textcolor{red}{\textbf{Dark Blue: 40\% believe it is important, but a lower priority.}}
\newline
- \textcolor{red}{\textbf{Grey: 13\% think it is not too important or should not be done.}}
\newline
- \textcolor{red}{\textbf{Black: 11\% don't know or have no opinion on the matter.}}
\\ \hline
\end{tabular}
}
\end{table*}
\begin{table*}[t]
\centering
\caption{The first figure illustrates an example of a more detailed response (i.e., including the exact chart type) of the GPT-4V model compared to the compact response of the Gemini-pro-vision model. The second figure demonstrates an example of a scenario where the GPT-4V model correctly identified that the chart in question is a \textit{`Pie chart'} and it does not have any x-axis or y-axis range, whereas the Gemini-pro-vision model fails to answer the question correctly. The correct text is marked in \textcolor{green}{\textbf{Green}} and erroneous text is marked in \textcolor{red}{\textbf{Red}}.}
\label{tab:l1-examples}
\scalebox{0.95}{
\begin{tabular}{p{0.5\linewidth}p{0.5\linewidth}}
\Xhline{4\arrayrulewidth}
\multicolumn{1}{l}{\textbf{Semantic Evaluation: Level - 1}} & \\ \hline \hline
\multicolumn{1}{p{0.5\linewidth}|}{\raisebox{-\height}{\includegraphics[width=0.5\textwidth]{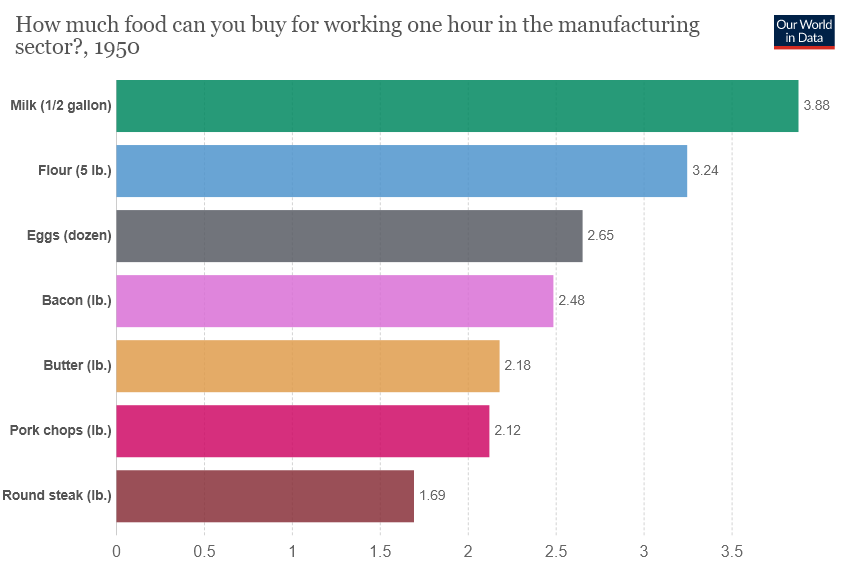}}} &
\textbf{Prompt:} What is the chart type in the input image?
\newline\textbf{Gemini-pro-vision Response:} The chart type in the input image is a \textit{bar chart}.
\newline\textbf{GPT-4V Response:} The chart in the image is a \textit{horizontal bar chart}. This type of chart is used to represent data with rectangular bars where the length of the bar is proportional to the value of the variable. In this bar chart, the bars represent the amount of various food items one could buy after working one hour in the manufacturing sector in the year 1950.
 \\ \midrule
\multicolumn{1}{p{0.5\linewidth}|}{\raisebox{-\height}{\includegraphics[width=0.5\textwidth]{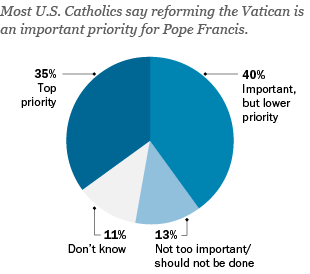}}} &
\textbf{Prompt:} What is the range of x-axis?
\newline\textbf{Gemini-pro-vision Response:} \textcolor{red}{\textbf{0 to 40.}}
\newline\textbf{GPT-4V Response:} The image you provided is a pie chart, and pie charts \textcolor{green}{\textbf{do not have an x-axis or y-axis}} as you would find in line graphs or bar charts. Instead, they display data in a circular format, with each slice of the pie representing a proportion of the whole. This particular pie chart is about the opinions of U.S. Catholics on Vatican Reform and does not include the concept of a range for an axis.
\\ \hline
\end{tabular}
}
\end{table*}
\begin{table*}[t]
\centering
\caption{The figure illustrates an example where both the GPT-4V and the Gemini-pro-vision model answer a question about \textit{Level - \textbf{2}} semantic content (i.e., maximum value) correctly. The correct text is marked in \textcolor{green}{\textbf{Green}}.}
\label{tab:l2-examples}
\scalebox{0.95}{
\begin{tabular}{p{0.5\linewidth}p{0.5\linewidth}}
\Xhline{4\arrayrulewidth}
\multicolumn{1}{l}{\textbf{Semantic Evaluation: Level - 2}} & \\ \hline \hline
\multicolumn{1}{p{0.5\linewidth}|}{\raisebox{-\height}{\includegraphics[width=0.5\textwidth]{emnlp2020-templates/Figures_llmeval_charts/l2-ex.png}}} &
\textbf{Prompt:} Identify the axis that contains a numerical range. What is the maximum value in that axis?
\newline\textbf{Gemini-pro-vision Response:} The y-axis contains a numerical range. The maximum value on the y-axis is \textcolor{green}{\textbf{18.7}}.
\newline\textbf{GPT-4V Response:} The axis that contains a numerical range in the image is the vertical (y-axis). It represents the "Multiple of annual gross earnings" for "Gross pension wealth." The maximum value on this axis is 18, which is the highest labeled horizontal line, but it does not match the value of the tallest bar. The tallest bar represents Luxembourg, with a value of 18.7, as indicated by the number above the bar. Hence, the actual maximum value depicted on the graph is \textcolor{green}{\textbf{18.7}}, even though the axis itself is only labeled up to 18.
\\ \hline
\end{tabular}
}
\end{table*}
\clearpage
\onecolumn
\begin{center}
\begin{longtable}{p{0.46\linewidth}p{0.46\linewidth}}
\caption{The first figure illustrates an example of the models' accurate response to a question about \textit{Level - \textbf{3}} semantic content presented in a chart (i.e., trend/pattern), and both the models could answer correctly as the trend was easy to follow and linearly increasing. However, when presented with a chart with multiple lines that represent a complex trend (i.e., a chart with many fluctuating lines), both the models fail to answer correctly, as depicted in the second and third figures. The correct text is marked in \textcolor{green}{\textbf{Green}} and erroneous text is marked in \textcolor{red}{\textbf{Red}}.}
\label{tab:l3-examples} \\
\Xhline{4\arrayrulewidth}
\multicolumn{1}{l}{\textbf{Semantic Evaluation: Level - 3}} &  \\ \hline \hline
\endfirsthead

\multicolumn{2}{c}%
{{\bfseries \tablename\ \thetable{} -- continued from previous page}} \\
\Xhline{4\arrayrulewidth}
\multicolumn{1}{l}{\textbf{Semantic Evaluation: Level - 3}} &  \\ \hline 
\endhead

\hline \multicolumn{2}{c}{{ -- Continued on next page --}} \\ \hline
\endfoot

\hline \hline
\endlastfoot
\multicolumn{1}{p{0.5\linewidth}}{\raisebox{-\height}{\includegraphics[width=0.5\textwidth]{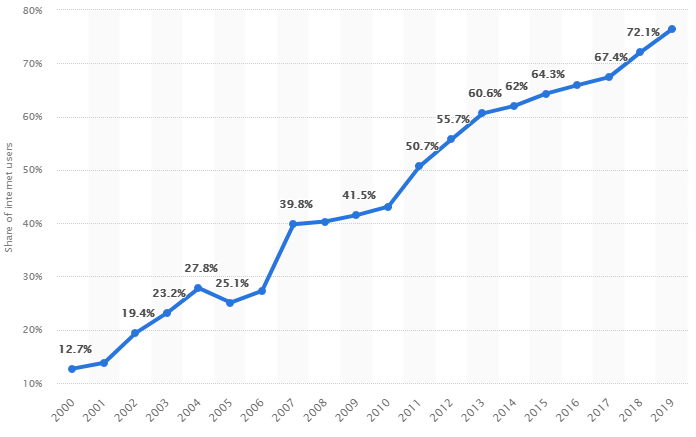}}} &
\textbf{Prompt:} Describe the trend that is visible in the chart and provide evidence for your conclusion.
\newline\textbf{Gemini-pro-vision Response:} The chart displays a \textcolor{green}{\textbf{clear upward trend}} over the period from 2000 to 2019. Starting at 12.7\% in 2000, the data point rises consistently throughout the years, reaching 72.1\% by 2019. This increase is steady with no evident declines, indicating a notable and persistent growth in the metric being measured, which is the ``Share of Internet users'' as seen on the y-axis. Every consecutive year shows an increase from the previous year. The x-axis, which represents time, clearly shows this is over a 20-year timespan.
The evidence for the conclusion is in the plotted data points themselves, each one being higher than the previous, without any dips or downward movements. The consistent increase in the percentage points reflects a robust and sustained growth in internet usage among the population being analyzed.
\newline\textbf{GPT-4V Response:}  The chart shows the share of internet users from 2000 to 2019. \textcolor{green}{\textbf{The trend is upward}}, with the share of internet users increasing from 12.7\% in 2000 to 76.1\% in 2019. The evidence for this conclusion is that the line on the chart is going up.
 \\ \midrule

\multicolumn{1}{p{0.5\linewidth}}{\raisebox{-\height}{\includegraphics[width=0.5\textwidth]{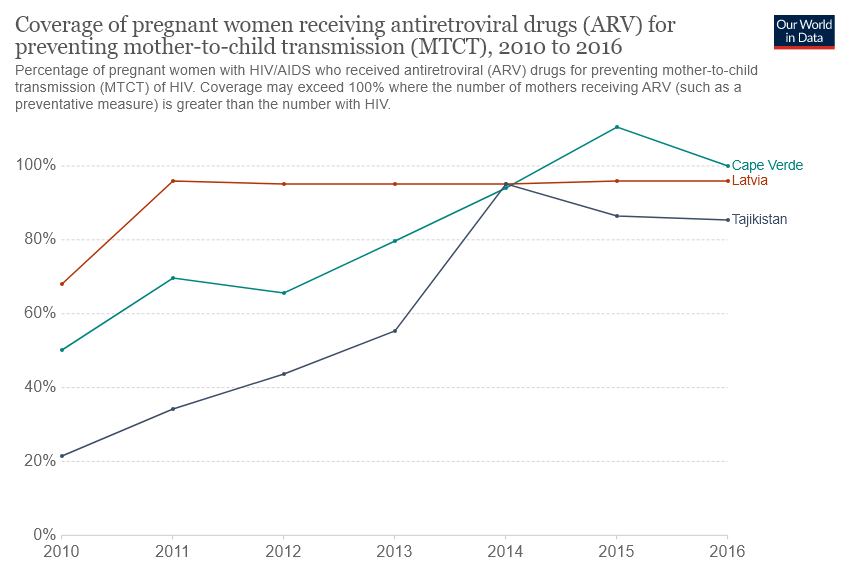}}} &
\textbf{Prompt:} Describe the trend that is visible in the chart and provide evidence for your conclusion.
\newline\textbf{Gemini-pro-vision Response:} The chart shows the coverage of pregnant women receiving antiretroviral drugs (ARV) for preventing mother-to-child transmission (MTCT) of HIV. \textcolor{red}{\textbf{ The coverage in Cape Verde increased from 50\% in 2010 to 98\% in 2016. The coverage in Latvia increased from 22\% in 2011 to 88\% in 2016. The coverage in Tajikistan increased from 58\% in 2013 to 95\% in 2016.}}
\newline\textbf{GPT-4V Response:} The chart presents data on the coverage of pregnant women receiving antiretroviral drugs (ARV) for preventing mother-to-child transmission (MTCT) of HIV in three different countries—Cape Verde, Latvia, and Tajikistan—from 2010 to 2016. A clear trend from this chart is that coverage increased in all three countries over the given time period.
\newline 1. \textcolor{red}{\textbf{Cape Verde shows a consistently high coverage close to or at 100\% throughout the period, with slight fluctuations but generally maintaining a plateau.}}
\newline 2. \textcolor{red}{\textbf{Latvia presents a dramatic increase starting from just above 20\% in 2010 to reaching approximately 80\% by 2016.}}
\newline 3. \textcolor{red}{\textbf{Tajikistan also shows considerable improvement, with the coverage starting at around 40\% in 2010 and ending up just below 80\% in 2016.}}
The evidence for these trends is in the trajectory of the lines for each country, which show upward slopes over the years, reflecting an increasing percentage of pregnant women receiving ARV drugs to prevent MTCT of HIV. Each point on the lines represents the data recorded for the corresponding year, clearly illustrating the progress each country has made throughout these years.
\\ \midrule
\multicolumn{1}{p{0.5\linewidth}}{\raisebox{-\height}{\includegraphics[width=0.5\textwidth]{emnlp2020-templates/Figures_llmeval_charts/l3-ex-3.png}}} &
\textbf{Prompt:} Describe the trend that is visible in the chart and provide evidence for your conclusion.
\newline\textbf{Gemini-pro-vision Response:} The chart shows that the saving rate in Belgium and Mexico has been on \textcolor{red}{\textbf{a downward trend since the early 1990s. While Belgium's saving rate has remained relatively stable since the mid-2000s, }} Mexico's saving rate has seen a slight recovery since the early 2000s.
\newline\textbf{GPT-4V Response:} The chart you provided shows the saving rates as a percentage of GDP for Belgium and Mexico from 1984 to 2006. Here's the trend based on the information presented:
For Belgium:
\newline
\textcolor{red}{\textbf{- The saving rate starts off at around 10\% in 1984.
\newline - There's a slight decrease until a sharp drop around 1986, where it reaches just above 3\%, the lowest point on the chart for Belgium.
\newline - A steady increase follows until around 1989, where the saving rate exceeds 10\% again.
\newline - From 1990 to the end of the chart, the saving rate fluctuates but generally maintains within the range of roughly 9\% to slightly above 11\%, without any drastic changes. For Mexico:
\newline - The saving rate is shown starting from 1990 on this chart (indicating no data provided or not tracked for the prior years).
\newline - It starts out at around 6\% in 1990.
\newline - There's a rather volatile pattern with significant ups and downs until 1995, where it dips to the lowest point, close to 4\%.
\newline - From this point on, there is a relatively steady upward trend, with some fluctuations but a clear overall increase in the saving rate.
\newline - This ends with a sharp increase between 2004 and 2006, where the saving rate jumps to around 6\% to just under 11\%.}}
Evidence for these trends comes from observing the lines on the chart and noting the changes in direction
\\ 
\end{longtable}
\end{center}
\clearpage
\twocolumn
\begin{table*}[t]
\centering
\caption{The figure illustrates examples where the Gemini-pro-vision and the GPT-4V models could produce domain-specific information pertaining to semantic content \textit{Level - \textbf{4}}. The text containing domain-specific information is marked in \textcolor{blue}{\textbf{Blue}}.}
\label{tab:l4-examples}
\scalebox{0.95}{
\begin{tabular}{p{0.5\linewidth}p{0.5\linewidth}}
\Xhline{4\arrayrulewidth}
\multicolumn{1}{l}{\textbf{Semantic Evaluation: Level - 4}} & \\ \hline \hline
\multicolumn{1}{p{0.5\linewidth}|}{\raisebox{-\height}{\includegraphics[width=0.5\textwidth]{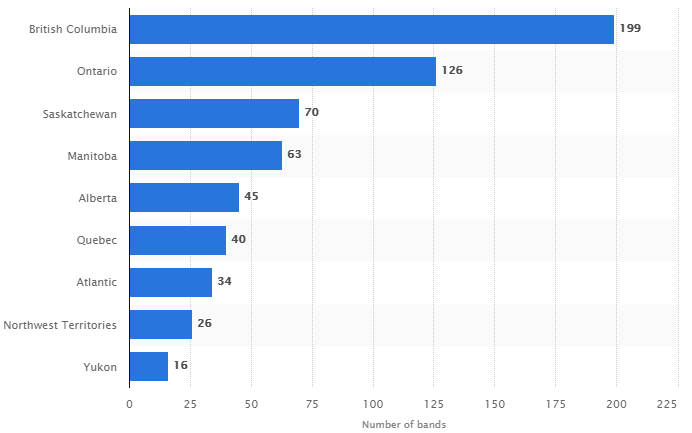}}} &
\textbf{Prompt:} Analyze the chart given in the input in one paragraph.
\newline\textbf{Gemini-pro-vision Response:} \textcolor{blue}{\textbf{ The chart shows the number of bands in each province and territory of Canada.}} As of 2021, British Columbia has the most bands, with 199, followed by Ontario with 126, and Saskatchewan with 70. The Northwest Territories has the least bands, with 26.
\newline\textbf{GPT-4V Response:} \textcolor{blue}{\textbf{The chart presented is a horizontal bar chart showing the number of bands in various regions or provinces of a country, which appears to be Canada given the regions listed.}} British Columbia leads with the highest number at 199 bands, followed by Ontario with 126 bands. These two regions have significantly more bands compared to the others. Saskatchewan comes in third with 70 bands, while Manitoba has a close 63. Alberta and Quebec show a mid-range number of bands, with 45 and 40 respectively. The Atlantic region is represented with 34 bands, and the Northwest Territories have 26. Yukon has the fewest bands, represented by a count of 16. \textcolor{blue}{\textbf{Overall, the chart provides a clear visualization of a disparity in the number of bands among the regions, hinting at a possible correlation between population, cultural factors, or policies that influence the number of music bands in these areas.}}
 \\ \midrule
\multicolumn{1}{p{0.5\linewidth}|}{\raisebox{-\height}{\includegraphics[width=0.5\textwidth]{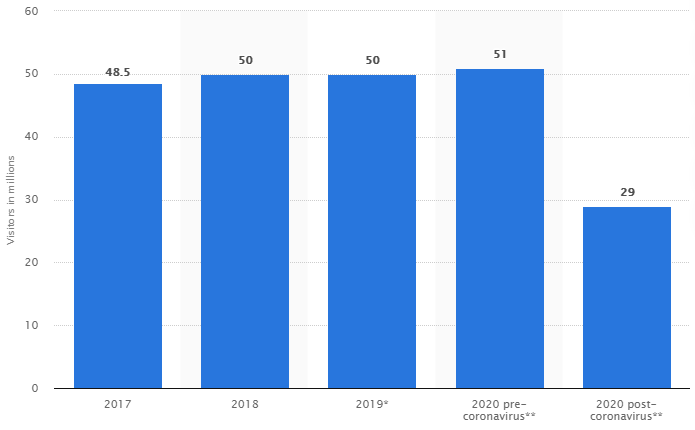}}} &
\textbf{Prompt:} Analyze the chart given in the input in one paragraph.
\newline\textbf{Gemini-pro-vision Response:} The chart shows the number of visitors in millions from 2017 to 2020. The number of visitors increased from 48.5 million in 2017 to 50 million in 2018 and remained the same in 2019. \textcolor{blue}{\textbf{However, the number of visitors decreased to 51 million in 2020 due to the COVID-19 pandemic.}} The number of visitors decreased even further to 29 million in 2020 post-coronavirus.
\newline\textbf{GPT-4V Response:} The chart is a bar graph representing the number of visitors in millions over a period from 2017 through 2020, where 2020 is split into ``pre-coronavirus'' and "post-coronavirus" periods, indicating the impact of the COVID-19 pandemic on visitor numbers. From 2017 to 2020 pre-coronavirus, there is an upward trend in visitor numbers, starting at approximately 48.5 million in 2017 and peaking at 51 million just before the coronavirus impact. \textcolor{blue}{\textbf{Notably, in the post-coronavirus period of 2020, there is a significant drop to 29 million visitors, reflecting the drastic reduction likely due to travel restrictions, lockdowns, and other measures put in place to combat the spread of COVID-19.}}
\\ \hline
\end{tabular}
}
\end{table*}

\end{document}